\documentclass[english]{article}
\usepackage{proceed2e}

\usepackage{times}				%

\usepackage{natbib}
\usepackage[colorlinks,citecolor=blue,urlcolor=blue,linkcolor=blue]{hyperref}
\usepackage{float}				%
\usepackage{caption}
\usepackage{subcaption}			%

\usepackage{algorithm}
\usepackage{algorithmicx}
\usepackage[noend]{algpseudocode} %
\newcommand{\algcomment}[1]{\Comment{\textit{#1}}}
\algnewcommand{\LineComment}[1]{\(\triangleright\) \textit{#1}}

\usepackage{multicol}

\usepackage{amsmath}					%
\usepackage{amsfonts}					%
\usepackage{bm}							%
\usepackage{mathtools}					%
\def\Reals{\mathbb{R}}					%
\def\bphi{\bm{\phi}}					%
\renewcommand{\d}{\,\mathrm{d}} 		%
\newcommand{\IE}{\mathbb{E}}  			%
\newcommand{\IP}{\mathbb{P}}  			%
\newcommand{\Exp}{\operatorname{Exp}}	%
\newcommand{\chol}{\operatorname{chol}}	%
\newcommand{\bPhi}{\mathbf{\Phi}}
\newcommand{\bw}{\mathbf{w}}
\newcommand{\bx}{\mathbf{x}}
\newcommand{\by}{\mathbf{y}}
\newcommand{\calA}{\mathcal{A}}
\newcommand{\calC}{\mathcal{C}}

\newcommand{\calX}{\mathcal{X}}

\usepackage{enumitem}
\setlist{nosep}

\usepackage{amsthm}
\theoremstyle{plain}
\newtheorem{proposition}{Proposition}
\newtheorem{lemma}{Lemma}
\theoremstyle{definition}
\newtheorem{definition}{Definition}

\title{The Mondrian Kernel}

\author{
Matej Balog\thanks{\ Also affiliated with Max-Planck Institute for Intelligent Systems, T\"ubingen, Germany.
} \\
Department of Engineering \\
University of Cambridge
\And
Balaji Lakshminarayanan\ \ \\
Gatsby Unit\\
University College London  
\AND
\ \ Zoubin Ghahramani \ \ \\
Department of Engineering \\
University of Cambridge
\And
Daniel M. Roy\\
Department of Statistical Sciences\\
University of Toronto
\And
Yee Whye Teh\\
Department of Statistics \\
University of Oxford
}

\begin{document}

\maketitle

\begin{abstract}
We introduce the Mondrian kernel, a fast \emph{random feature} approximation to the Laplace kernel. It is suitable for both batch and online learning, and admits a fast kernel-width-selection procedure as the random features can be re-used efficiently for all kernel widths. The features are constructed by sampling trees via a Mondrian process \citep{RT09}, and we highlight the connection to Mondrian forests \citep{MF}, where trees are also sampled via a Mondrian process, but fit independently. This link provides a new insight into the relationship between kernel methods and random forests.
\end{abstract}

\section{INTRODUCTION}

Kernel methods such as support vector machines and Gaussian processes are very popular in machine learning. While early work relied on dual optimization, recent large-scale kernel methods focus on the primal optimization problem where the input data are mapped to a finite-dimensional feature space and the weights are learned using fast \mbox{linear} optimization techniques, e.g., stochastic gradient descent. \citet{rahimi2007random} proposed to approximate shift-invariant kernels by mapping the inputs to so-called \emph{random features}, constructed so that the inner product of two mapped data points approximates the kernel evaluated at those two points (which is the inner product in the feature space corresponding to the kernel). \citet{rahimi2007random} proposed two random feature construction schemes: \emph{random Fourier features}, where data points are projected onto random vectors drawn from the Fourier transform of the kernel and then passed through suitable non-linearities; and \emph{random binning}, where the input space is partitioned by a random regular grid into bins and data points are mapped to indicator vectors identifying which bins they end up in. Both of these approaches require specifying the kernel hyperparameters in advance, so that the appropriate distribution is used for sampling the random vectors or random grids, respectively. However, a suitable kernel width (length-scale) is often not known a priori and is found by cross-validation, or, where available, marginal likelihood optimization. In practice, this entails constructing a new feature space and training a linear learner from scratch for each kernel width, which is computationally expensive. Using a suitable kernel width is often more important than the choice of kernel type \citep{scholkopf_learning_2001}, so a fast kernel width selection method is desirable.

We describe a connection between the Laplace kernel and the Mondrian process \citep{RT09}, and leverage it to develop a random feature approximation to the Laplace kernel that addresses the kernel width selection problem. This approximation, which we call the \emph{Mondrian kernel}, involves random partitioning of data points using a Mondrian process, which can be efficiently reused for all kernel widths. The method preserves the nonparametric nature of kernel learning and is also suitable for online learning.

The Mondrian kernel reveals an interesting link between kernel methods and decision forests \citep{RF,DF}, another popular class of nonparametric methods for black-box prediction tasks. The Mondrian kernel resembles \emph{Mondrian forests}, a decision-forest variant introduced by \citet{MF}, where a Mondrian process is used as the randomization mechanism. The efficiently trainable Mondrian forests excel in the online setting, where their distribution is identical to the corresponding batch Mondrian forest, and have been successfully applied to both classification and regression \citep{MF, MFreg}. Mondrian forests and the Mondrian kernel both lead to randomized, non-linear learning algorithms whose randomness stems from a Mondrian process. The former fits parameters corresponding to different Mondrian trees independently, while the latter fits them jointly. We compare these methods theoretically and thus establish a novel connection between the Laplace kernel and Mondrian forests via the Mondrian kernel.

The contributions of this paper are:
\begin{itemize}
\item a review of the Mondrian process using the simple notion of competing exponential clocks (Section~\ref{sec:TheMondrianProcess});
\item a novel connection between the Mondrian process and the Laplace kernel (Section~\ref{sec:TheMondrianKernel}), yielding a fast approximation to learning with the Laplace kernel;
\item an efficient procedure for learning the kernel width from data (Section~\ref{sec:LengthscaleLearning}); and
\item a comparison between Mondrian kernel and Mondrian forest that provides another connection between kernel learning and random forests (Section~\ref{sec:ForestConnection}).
\end{itemize}

\section{MONDRIAN PROCESS}
\label{sec:TheMondrianProcess}

For completeness, we review the Mondrian process \citep[Chapter 5]{RT09,RoyThesis}. Although simple and perhaps well known to experts, our exposition through competing exponential clocks has not explicitly appeared in this form in the literature. Readers familiar with the Mondrian process may skip this section on first reading.

\subsection{TERMINOLOGY}

An \emph{axis-aligned box} $\calX = \calX_1 \times \cdots \times \calX_D \subseteq \Reals^D$ is a Cartesian product of $D$ bounded intervals $\calX_d \subseteq \Reals$. Their total length $|\calX_1| + \cdots + |\calX_D|$ is the \emph{linear dimension of} $\calX$. A \emph{guillotine partition} of $\calX$ is a hierarchical partitioning of $\calX$ using axis-aligned cuts. Such a partition can be naturally represented using a strictly binary tree.

An \emph{exponential clock with rate} $r$ takes a random time $T \sim \Exp(r)$ to ring after being started, where $\Exp(r)$ is the exponential distribution with rate (inverse mean) $r$. The notion of \emph{competing exponential clocks} refers to $D$ independent exponential clocks with rates $r_1, \ldots, r_D$, started at the same time. It can be shown that (1) the time until some clock rings has $\Exp( \sum r_d)$ distribution, (2) it is the $d$-th clock with probability proportional to $r_d$, and (3) once a clock rings, the remaining $D - 1$ clocks continue to run independently with their original distributions.

\subsection{GENERATIVE PROCESS}
\label{sec:MondrianGenerativeProcess}

The Mondrian process on an axis-aligned box $\calX \subseteq \Reals^D$ is a time-indexed stochastic process taking values in guillotine-partitions of $\calX$. It starts at time $0$ with the trivial partition of $\calX$ (no cuts) and as time progresses, new axis-aligned cuts randomly appear, hierarchically splitting $\calX$ into more and more refined partitions. The process can be stopped at a lifetime $\lambda \in [0, \infty)$, which amounts to ignoring any cuts that would appear after time $\lambda$.

To describe the distribution of times and locations of new cuts as time progresses, we associate an independent exponential clock with rate $|\calX_d|$ to each dimension $d$ of $\calX$. Let $T$ be the first time when a clock rings and let $d$ be the dimension of that clock. If $T > \lambda$ then this process terminates. Otherwise, a point $a$ is chosen uniformly at random from $\calX_d$ and $\calX$ is split into $\calX^{<} = \{ \bx \in \calX \mid x_d < a \}$ and $\calX^{>} = \{ \bx \in \calX \mid x_d > a \}$ by a hyperplane in dimension $d$ that is perpendicular to $\calX_d$ at point $a$. After making this first cut, the remaining $D - 1$ clocks are discarded and the generative process restarts recursively and \emph{independently} on $\calX^{<}$ and $\calX^{>}$. However, those processes start at time $T$ rather than $0$ and thus have less time left until the lifetime $\lambda$ is reached.

The specification of the generative process on $\calX$ is now complete. %
Due to the properties of competing exponential clocks, 
the time until the first cut appears in $\calX$ has exponential distribution with rate equal to the linear dimension of $\calX$ and the dimension $d$ in which the cut is made is chosen proportional to $|\calX_d|$. This confirms equivalence of our generative process to the one proposed by \citet{RT09}. Finally, we note that a.s.~the Mondrian process does not explode, i.e., for every lifetime $\lambda \in [0, \infty)$, the process generates finitely many cuts with probability $1$ \citep{RoyThesis}.

\begin{figure}[tb]
\centering
\begin{subfigure}[t]{0.20\textwidth}
\centering
\includegraphics[width=\linewidth]{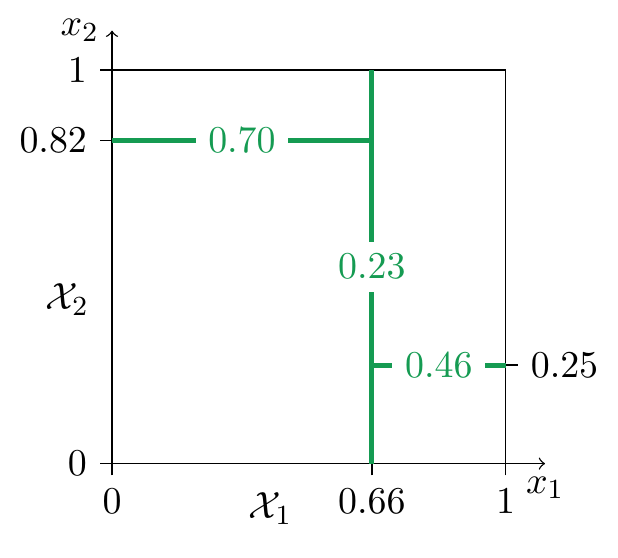}
\caption{}
\end{subfigure}%
~
\begin{subfigure}[t]{0.28\textwidth}
\centering
\includegraphics[width=\linewidth]{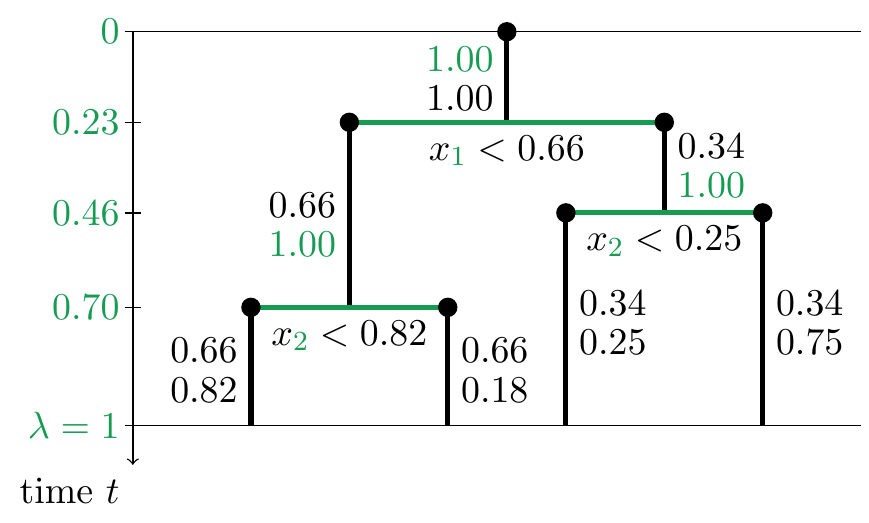}
\caption{}
\end{subfigure}%
\caption{(a) Sample of a Mondrian process on the axis-aligned box $\calX = [0, 1] \times [0, 1] \subseteq \Reals^2$ with lifetime $\lambda = 1.0$. Numbers on the cuts (shown in green) indicate the times when they appeared. The first cut appeared at time $T = 0.23$, in dimension $d = 1$, at location $a = 0.66 \in \calX_1$. (b) Representing the Mondrian sample as a strictly binary tree, with new nodes (shown as circles) appearing as time (y-axis) progresses. The two numbers below each node show the rates of the two exponential clocks competing to split that node, with the winning clock's rate shown in green.}
\label{fig:MondrianSample}
\end{figure}

\subsection{PROJECTIVITY}

\begin{figure}[t!]
\vspace{-0.5em}
\begin{subfigure}[t]{0.25\textwidth}
\hspace{0.5em}
\includegraphics[scale=0.74]{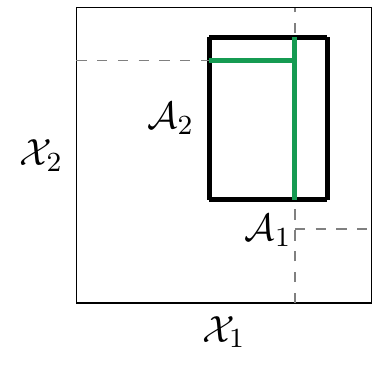}
\caption{}
\label{fig:ConsistencySubfigure1}
\end{subfigure}%
~ 
\begin{subfigure}[t]{0.25\textwidth}
\hspace{-0.1em}
\includegraphics[scale=0.74]{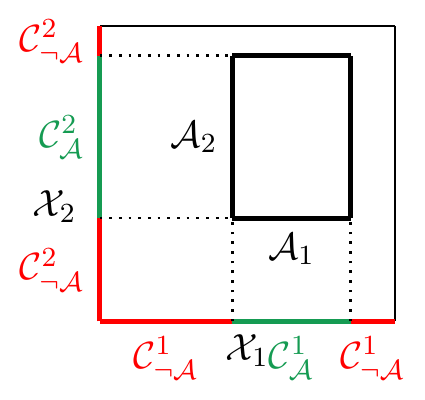}
\caption{}
\label{fig:ConsistencySubfigure2}
\end{subfigure}%

\begin{subfigure}[t]{0.25\textwidth}
\hspace{0.5em}
\includegraphics[scale=0.74]{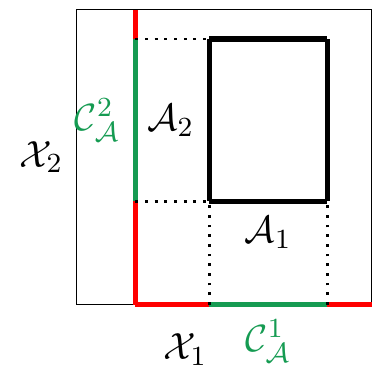}
\caption{}
\label{fig:ConsistencySubfigure3}
\end{subfigure}%
~
\begin{subfigure}[t]{0.25\textwidth}
\hspace{0.9em}
\includegraphics[scale=0.74]{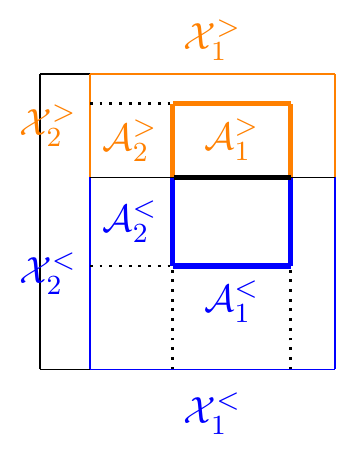}
\caption{}
\label{fig:ConsistencySubfigure4}
\end{subfigure}
\caption{(a) A Mondrian process running on $\calX = \calX_1 \times \calX_2$ generates cuts (dashed lines), some of which intersect $\calA = \calA_1 \times \calA_2$ (green lines) and thus induces a random partition of $\calA$. (b) Representing the first cut distribution using $2D = 4$ competing exponential clocks: in each dimension $d$, clock $\calC^d_{\calA}$ corresponds to the region where making a cut splits $\calA$ (shown in green) and clock $\calC^d_{\lnot \calA}$ to the (disconnected) region where making a cut misses $\calA$ (shown in red). (c) Cut outside $\calA$: reusing the clocks $\calC^1_{\calA}$, $\calC^2_{\calA}$ on the side of the cut containing $\calA$. (d) Cut inside $\calA$ (shown in black): the argument proceeds by induction on both sides.}
\label{fig:ConsistencyIntuitionFigure}
\end{figure}

If a Mondrian process runs on $\calX$, what distribution of random partitions does it induce on an axis-aligned subbox $\calA \subseteq \calX$? (See Figure~\ref{fig:ConsistencySubfigure1} for an illustration in $D = 2$ dimensions.) The Mondrian process was constructed so that the answer is the Mondrian process itself \citep{RoyThesis}. Here we explain this projectivity property using the notion of competing exponential clocks. To argue that the resulting process on $\calA$ is indeed a Mondrian process, we show that the process running on $\calX$ generates cuts in $\calA$ in the same way as a Mondrian process running directly on $\calA$ would.

Recall that each dimension $d$ of $\calX$ is associated with an exponential clock with rate $|\calX_d|$ and if it rings first, the cut location is chosen uniformly at random from $\calX_d$. This procedure can be equivalently represented using two competing clocks for each dimension (rather than just one):
\begin{itemize}
\item Clock $\calC^d_{\calA}$ with rate $|\calA_d|$. If this clock rings first, the cut location is chosen uniformly at random from $\calA_d$.
\item Clock $\calC^d_{\lnot \calA}$ with rate $|\calX_d| - |\calA_d|$. If it rings first, the cut location is sampled uniformly from $\calX_d \setminus \calA_d$.
\end{itemize}

(See Figure~\ref{fig:ConsistencySubfigure2}.) Note that the clocks $\calC^1_{\calA}, \ldots, \calC^D_{\calA}$ represent the same cut distribution as a Mondrian process running on $\calA$ would. If a clock $\calC^d_{\lnot \calA}$ rings first, a cut is made outside of $\calA$ and all of $\calA$ remains on one side of this cut. None of the clocks $\calC^d_{\calA}$ have rung in that case and would usually be discarded and replaced with fresh clocks of identical rates, but by property (3) of competing exponential clocks, we can equivalently reuse these clocks (let them run) on the side of the cut containing $\calA$. (Figure~\ref{fig:ConsistencySubfigure3} shows a cut in dimension $d=1$ that misses $\calA$ and the reused clocks $\calC^d_{\calA}$). Hence, cuts outside $\calA$ do not affect the distribution of the first cut crossing $\calA$, and this distribution is the same as if a Mondrian process were running just on $\calA$. When a cut is made within $\calA$ (see Figure~\ref{fig:ConsistencySubfigure4}), the process continues on both sides recursively and our argument proceeds inductively, confirming that the Mondrian process on $\calX$ generates cuts in $\calA$ in the same way as a Mondrian process on $\calA$ would.

\subsection{MONDRIAN PROCESS ON $\Reals^D$}
\label{sec:MondrianProcessUnbounded}

The Mondrian process on $\Reals^D$ is defined implicitly as a time-indexed stochastic process such that its restriction to any axis-aligned box $\calX \subseteq \Reals^D$ is a Mondrian process as defined in section~\ref{sec:MondrianGenerativeProcess}. 
Fortunately, this infinite-dimensional object can be compactly represented
 by instantiating the Mondrian process only in regions where we have observed data. As we observe new data points, the Mondrian sample can be extended using the  \emph{conditional Mondrian} algorithm \citep{RT09}, a simple and fast sampling procedure for extending a Mondrian sample in an axis-aligned box $\calA$ to a larger axis-aligned box $\calX \supseteq \calA$. 
The conditional Mondrian is useful for online learning and prediction, as it can be used to extend Mondrian samples to (yet) unobserved parts of the input space \citep{MF}.

\section{MONDRIAN KERNEL}
\label{sec:TheMondrianKernel}

For concreteness, our running example will be regression: the problem of learning a function $f : \Reals^D \to \Reals$ from a set of $N$ training examples $(\mathbf{x}_1, y_1), \ldots, (\mathbf{x}_N, y_N)$. However, the Mondrian kernel applies equally well to classification, or any other learning task.

Learning with kernels involves choosing a kernel function $k : \Reals^D \times \Reals^D \to \Reals$ to act as a similarity measure between input data points. Evaluating $k(\cdot, \cdot)$ on all pairs of $N$ data points takes $\Omega(N^2)$ operations, with some models also requiring a $\Theta(N^3)$ operation on an $N \times N$ kernel matrix. This generally makes exact kernel methods unsuitable for large-scale learning. \citet{rahimi2007random} proposed a fast approximation through a randomized construction of a low-dimensional feature map $\phi : \Reals^D \to \Reals^C$ such that
\begin{equation*}
\forall{\bx, \bx' \in \Reals^D}
\hspace{2em}
k(\bx, \bx')
\approx \phi(\bx)^T \phi(\bx')
\end{equation*}
and then using a linear learning method in the feature space $\Reals^C$ implied by $\phi$. For example, linear regression $\by\approx\bPhi\bw$, where $\bPhi \in \Reals^{N \times C}$ is the feature matrix with $n$-th row $\phi(\bx_n)^T$, is solvable exactly in time linear in $N$.
In general, the primal problem also lends itself naturally to stochastic gradient descent approaches for learning $\bw$.

We use the Mondrian process to construct a randomized feature map for the (isotropic) \emph{Laplace kernel}:
\begin{equation*}
k(\bx, \bx')
= \exp(- \lambda \| \bx - \bx' \|_1)
= \exp(- \lambda \sum_{d = 1}^D |x_d - x'_d|).
\end{equation*}
Here $\lambda \geq 0$ is the inverse kernel width (inverse length-scale), which we call the \emph{lifetime} parameter of the kernel. We use a non-standard parametrization as this lifetime parameter will be linked to the Mondrian process lifetime.

\subsection{MONDRIAN KERNEL}
\label{sec:MondrianKernel}

Consider the following randomized construction of a feature map $\phi : \Reals^D \to \Reals^C$:
\begin{enumerate}
\item Sample a partition of $\Reals^D$ via a Mondrian process on $\Reals^D$ with lifetime $\lambda$. Label the cells of the generated partition by $1, 2, \ldots$ in arbitrary order.
\item To encode a data point $\bx \in \Reals^D$, look up the label $c$ of the partion cell $\bx$ falls into and set $\phi(\bx)$ to be the (column) indicator vector that has a single non-zero entry at position $c$, equal to $1$.
\end{enumerate}
The Mondrian process on $\Reals^D$ generates infinitely many partition cells and cannot be stored in memory, but projectivity comes to the rescue. As we only ever need to evaluate $\phi$ on finitely many data points, it suffices to run the Mondrian on the smallest axis-aligned box containing all these points. Also, we only label partition cells containing at least one data point, in effect removing features that would be $0$ for all our data points. Then, the dimensionality $C$ of $\phi$ equals the number of non-empty partiton cells and each data point has a single non-zero feature, equal to $1$.

\begin{figure}[H]
\centering
\begin{minipage}{.23\textwidth}
\centering
\includegraphics[scale=0.7]{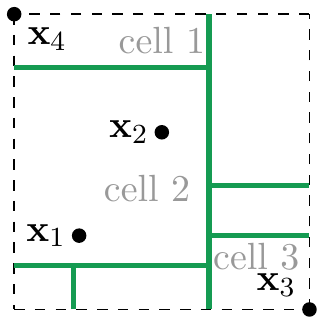}
\end{minipage}
~
\begin{minipage}{.23\textwidth}
\centering
\begin{tabular}{ c | c c c }
  $\mathbf{x}$ & $\phi(\mathbf{x})^T$ & \\\hline
  $\mathbf{x}_1$ & [0 1 0] \\
  $\mathbf{x}_2$ & [0 1 0] \\
  $\mathbf{x}_3$ & [0 0 1] \\
  $\mathbf{x}_4$ & [1 0 0]
\end{tabular}
\end{minipage}
\caption{Feature expansions of $4$ data points in $\Reals^2$.}
\label{fig:MondrianFeaturesExample}
\end{figure}

However, note that the set of points on which the feature map $\phi$ is evaluated need not be known in advance and can even grow in an online fashion. Indeed, the conditional Mondrian algorithm discussed in section~\ref{sec:MondrianProcessUnbounded} allows us to extend Mondrian samples to larger boxes as necessary, and we can increase the dimensionality of $\phi$ whenever a data point is added to a previously empty partition cell.

This feature map $\phi$ induces a kernel
\begin{align}
k_1(\bx, \bx') &:=
  \phi(\bx)^T \phi(\bx') \nonumber
\\ &=
\begin{dcases}
  1 & \text{ if } \bx, \bx' \text{ in same partition cell } \\
  0 & \text{ otherwise }
\end{dcases}
\label{eq:k1}
\end{align}
which we call a \emph{Mondrian kernel} of order $1$.

Instead of using a single Mondrian sample (partition), we can use $M$ independent samples and construct a feature map $\phi$ by concatenating and normalizing the feature maps $\phi^{(1)}, \ldots, \phi^{(M)}$ obtained from each individual sample as above:
\begin{equation}
\phi(\bx)
:= \frac{1}{\sqrt{M}} \left[ \phi^{(1)}(\bx)^T \;\;\cdots\;\; \phi^{(M)}(\bx)^T \right]^T.
\label{eq:PhiConcat}
\end{equation}
This feature expansion is \emph{sparse}: every data point has exactly $M$ non-zero features. The corresponding kernel, which we call a \emph{Mondrian kernel of order} $M$, is
\begin{align*}
  k_M(\bx, \bx') &:=
  \phi(\bx)^T \phi(\bx')
\\ &=
  \frac{1}{M} \sum_{m = 1}^M \phi^{(m)}(\bx)^T \phi^{(m)}(\bx').
\end{align*}
This is the empirical frequency with which points $\bx$ and $\bx'$ end up in the same partition cell of a Mondrian sample.

\begin{algorithm}[H]
\caption{Mondrian kernel}
\begin{algorithmic}[1]
\For{$m = 1$ \textbf{to} $M$}
\State construct feature map $\phi^{(m)}$ 
\algcomment{section~\ref{sec:MondrianKernel}}
\EndFor
\State join and rescale $\phi^{(1)}, \ldots, \phi^{(M)}$ into $\phi$
\algcomment{equation (\ref{eq:PhiConcat})}
\State map data $\mathbf{X}$ to feature representations $\bPhi$ using $\phi$
\State use linear learning method on $\bPhi$
\end{algorithmic}
\end{algorithm}

\subsection{MONDRIAN--LAPLACE LINK}
\label{sec:MondrianLaplacianLink}

By independence of the $M$ Mondrian samples, a.s.
\begin{equation*}
\lim_{M \to \infty} k_M(\bx, \bx')
= \IE\left[ \phi^{(1)}(\bx)^T \phi^{(1)}(\bx') \right]
= \IE\left[ k_1(\bx, \bx') \right]
\end{equation*}
with convergence at the standard rate $\mathcal{O}_p(M^{-1/2})$. We thus define the \emph{Mondrian kernel of order} $\infty$ as
\begin{equation*}
k_{\infty}(\bx, \bx') := \IE[ k_1(\bx, \bx')].
\end{equation*}

\begin{proposition}[Mondrian-Laplace link]
The Mondrian kernel of order $\infty$ coincides with the Laplace kernel.
\begin{proof}
As $k_1(\bx, \bx')$ (defined in (\ref{eq:k1})) is a binary random variable, $k_{\infty}(\bx, \bx')$ equals the probability that $\bx$ and $\bx'$ fall into the same partition cell of a Mondrian sample, which is equivalent to the sample having no cut in the minimal axis-aligned box spanned by $\bx$ and $\bx'$.

\begin{figure}[H]
\centering
\includegraphics[width=0.75\columnwidth]{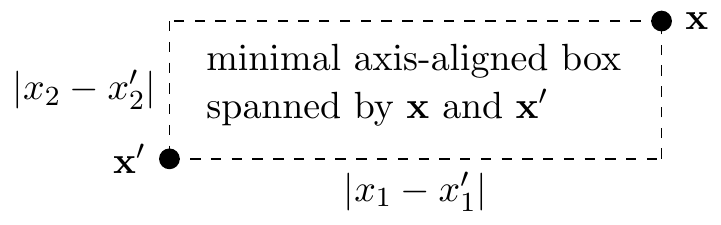}
\label{fig:MondrianLaplacianProofIllustration}
\vspace{-0.25em}
\end{figure}

By projectivity, this probability is the same as the probability of not observing any cuts in a Mondrian process with lifetime $\lambda$ running on just this minimal box. Noting that the linear dimension of this box is $\| \bx - \bx' \|_1$, we obtain
\begin{align*}
k_{\infty}(\bx, \bx') & =
  \IP(\text{no cut between } \bx, \bx' \text{ until time } \lambda )
\\ & =
  \IP\left( T > \lambda \right) \text{ where } T \sim \text{Exp}\left( \| \bx - \bx' \|_1 \right)
\\ & =
  e^{- \lambda \| \bx - \bx' \|_1}.
  \qedhere
\end{align*}
\end{proof}
\end{proposition}
Note that the lifetime (inverse width) $\lambda$ of the Laplace kernel corresponds to the lifetime of the Mondrian process used in the construction of the Mondrian kernel.

This link allows us to approximate the Laplace kernel with a Mondrian kernel $k_M$, which, unlike the Laplace kernel, admits a finite-dimensional feature expansion. The finite order $M$ trades off kernel approximation error and computational costs (indirectly through the complexity of $\phi$).

The following result confirms that the convergence of the Mondrian kernel approximation is exponentially fast in $M$ \emph{uniformly} on any fixed bounded input domain $\calX$.

\begin{proposition} For any bounded input domain $\calX \subseteq \Reals^D$ and $\delta > 0$, as $M \to \infty$,
\begin{align*}
&\;\;\;\;\;
  \IP\left[ \sup_{\bx, \bx' \in \calX} \left| k_M(\bx, \bx') - k_{\infty}(\bx, \bx') \right| > \delta \right]
\\ &=
  \mathcal{O}\left( M^{2/3} e^{- M \delta^2 / (12D + 2)} \right).
\end{align*}
\begin{proof}
Given in Supplement~\ref{sec:AppendixConvergenceBounds}.
\end{proof}
\end{proposition}

\section{FAST KERNEL WIDTH LEARNING}
\label{sec:LengthscaleLearning}

This section discusses the main advantage of our Mondrian approximation to the Laplace kernel: the efficient learning of kernel width from data. In particular, the approximation allows for efficient evaluation of all kernel lifetimes (inverse widths) $\lambda \in [0, \Lambda]$, where the terminal lifetime $\Lambda > 0$ need not be fixed a priori.

\subsection{FEATURE SPACE REUSAL}

\begin{figure}[t!]
\centering
\begin{minipage}{.23\textwidth}
\centering
\includegraphics[scale=0.7]{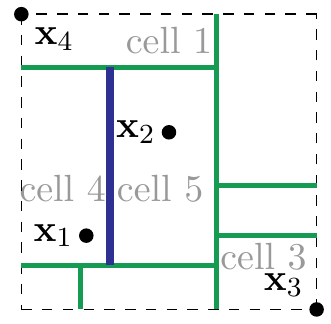}
\end{minipage}
~
\begin{minipage}{.23\textwidth}
\centering
\includegraphics[scale=0.9]{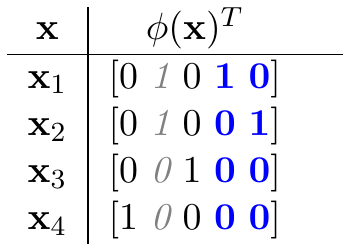}
\end{minipage}
\caption{A new cut (shown in thick blue) appeared, splitting cell $c = 2$ (cf.~Figure~\ref{fig:MondrianFeaturesExample}) into two new cells $c = 4$ and $c = 5$. The table shows the update to $\phi$, with the removed feature in gray italics and the two new features in bold blue.\vspace{-0.15em}}
\label{fig:MondrianFeaturesLifetimeUpdate}
\end{figure}

We make the following recollections from earlier sections:
\begin{itemize}
\item the Mondrian process runs through time, starting at time $0$ and only refining the generated partition as time progresses (cuts are never removed)
\item the Mondrian process with lifetime $\lambda$ is obtained by ignoring any cuts that would occur after time $\lambda$
\item the lifetime $\lambda$ of the Mondrian process used in constructing an explicit feature map $\phi$ for a Mondrian kernel corresponds to the lifetime (inverse width) of the Laplace kernel that it approximates
\end{itemize}
Running the Mondrian process from time $0$ to some terminal lifetime $\Lambda$ thus sweeps through feature spaces approximating all Laplace kernels with lifetimes $\lambda \in [0, \Lambda]$. More concretely, we start with $\lambda = 0$ and $\phi$ the feature map corresponding to $M$ trivial partitions, i.e., for any data point $\bx$, the vector $\phi(\bx)$ has length $M$ and all entries set to the normalizer $M^{-1/2}$. As we increase $\lambda$, at discrete time points new cuts appear in the $M$ Mondrian samples used in constructing $\phi$. Suppose that at some time $\lambda$, the partition cell corresponding to the $c$-th feature in $\phi$ is split into two by a new cut that first appeared at this time $\lambda$. We update the feature map $\phi$ by removing the $c$-th feature and appending two new features, one for each partition cell created by the split. See Figure~\ref{fig:MondrianFeaturesLifetimeUpdate} for an example with $M = 1$.

This procedure allows us to approximate all Laplace kernels with lifetimes $\lambda \in [0, \Lambda]$ without having to resample new feature spaces for each lifetime. The total computational cost is the same (up to a multiplicative constant) as of constructing a single feature space just for the terminal lifetime $\Lambda$. This is because a strictly binary tree with $C^{(m)}$ leaves (partition cells in the $m$-th Mondrian sample at time $\Lambda$) contains at most $C^{(m)} - 1$ internal nodes (features that had to be removed at some time point $\lambda < \Lambda$).

\subsection{LINEAR LEARNER RETRAINING}

Evaluating suitability of a lifetime (inverse kernel width) $\lambda$ requires training and evaluating a linear model in the feature space implied by $\phi$. This can also be done more efficiently than retraining a new model from scratch every time a new cut is added and $\phi$ updated. We discuss the example of ridge regression with exact solutions, and a general case of models trainable using gradient descent methods.

\subsubsection{Ridge regression}
\label{sec:LinearLearnerRetrainingRidgeRegression}

The MAP weights of the primal ridge regression problem are $\hat{\bw} = \mathbf{A}^{-1} \bPhi^T \by$, where $\mathbf{A} := (\bPhi^T \bPhi + \delta^2 \mathbf{I}_C)$ is the regularized feature covariance matrix and $\delta^2$ is the regularization hyperparameter. Instead of inverting $\mathbf{A}$, it is numerically more stable to work with its Cholesky factor $\chol(\mathbf{A})$ \citep{seeger_bayesian_2003}. Phrasing the problem as Bayesian linear regression with, say, observation noise variance $\sigma_{y}^2 = \delta^2$ and prior weights variance $\sigma_{w}^2 = 1$, we can also obtain the log marginal likelihood $\mathcal{L}(\lambda)$ of the form
\begin{equation*}
\mathcal{L}(\lambda)
= - \frac{\| \by - \bPhi \hat{\bw} \|_2^2}{2 \delta^2}
  - \frac{\| \hat{\bw} \|_2^2}{2} 
  - \frac{1}{2} \ln \det \mathbf{A}
  + \text{const},
\end{equation*}
where the dependence on $\lambda$ is implicit through $\phi$.

When a new cut appears in one of the $M$ Mondrian samples and $\phi$ is updated by deleting the $c$-th feature and appending two new ones, the corresponding update to the regularized feature covariance matrix $\mathbf{A}$ is to delete its $c$-th row and $c$-th column, and append two new rows and columns. Then both $\mathbf{A}^{-1}$ and $\chol(\mathbf{A})$ can be appropriately updated in $\mathcal{O}(C^2)$ time, faster than $\mathcal{O}(C^3)$ recomputation from scratch. Updating the Cholesky factor when the $c$-th row and column are removed is slightly involved but can be achieved by first permuting the rows and columns so that the ones to be removed are the last ones \citep{EPFL-REPORT-161468}, after which the Cholesky factor is updated by deleting its last row and column. If $C$ is the number of features at the terminal lifetime $\Lambda$, this $\mathcal{O}(C^2)$ update is performed $\mathcal{O}(C)$ times, for a total computational cost $\mathcal{O}(C^3)$. Note that performing the inversion or Cholesky factorization at just the terminal lifetime $\Lambda$ would have the same time complexity.

After updating $\mathbf{A}^{-1}$ or $\chol(\mathbf{A})$, the optimal weights $\hat{\bw}$ can be updated in $\mathcal{O}(C^2 + N)$ time and the determinant of $\mathbf{A}$ required for the marginal likelihood $\mathcal{L}(\lambda)$ can be obtained from $\chol(\mathbf{A})$ as the squared product of its diagonal elements in $\mathcal{O}(C)$ time. Exploiting sparsity of $\phi$, evaluating the model on $N_{\text{test}}$ data points takes $\mathcal{O}(N_{\text{test}} M)$ time.

Finally, we note that computing the marginal likelihood $\mathcal{L}(\lambda)$ for all $\lambda \in [0, \Lambda]$ and combining it with a prior $p(\lambda)$ supported on $[0, \Lambda]$ allows Bayesian inference over the kernel width $\lambda^{-1}$.
We refer to Supplement~\ref{sec:bayesian kernel width learning} for more details.

\subsubsection{Models trainable using gradient descent}

Consider a linear model trained using a gradient descent method. If (an approximation to) the optimal weight vector $\mathbf{w}$ is available and then $\phi$ is updated by removing the $c$-th feature and appending two new features, a natural way of reinitializing the weights for subsequent gradient descent iterations is to remove the $c$-th entry of $\mathbf{w}$ and append two new entries, both set to the removed value (as points in the split cell are partitioned into the two new cells, this preserves all model predictions). Note that we have the freedom of choosing the number of gradient descent iterations after each cut is added, and we can opt to only evaluate the model (on a validation set, say) at several $\lambda$ values on the first pass through $[0, \Lambda]$.
One iteration of stochastic gradient descent takes $O(M)$ time thanks to sparsity of $\phi$.

This efficient kernel width selection procedure can be especially useful with models where hyperparameters cannot be tweaked by marginal likelihood optimization (e.g., SVM).

\section{ONLINE LEARNING}
\label{sec:OnlineLearning}

In this section, we describe how the Mondrian kernel can be used for online learning. When a new data point $\bx_{N + 1} \in \Reals^D$ arrives, incorporating it into $M$ existing Mondrian samples (using the conditional Mondrian algorithm discussed in section~\ref{sec:MondrianProcessUnbounded}) can create $0 \leq k \leq M$ new non-empty partition cells, increasing the dimensionality of the feature map $\phi$. We set the new features to $0$ for all previous data points $\bx_1, \ldots, \bx_N$.

In our running example of ridge regression, exact primal updates can again be carried out efficiently. The inverse $\mathbf{A}^{-1}$ or Cholesky factor $\chol(\mathbf{A})$ of the regularized feature covariance matrix $\mathbf{A}$ can be updated in two steps:
\begin{enumerate}
\item extend $\mathbf{A}^{-1}$ or $\chol(\mathbf{A})$ to incorporate the $k$ new features (set to $0$ for all existing data points) in $\mathcal{O}(C^2)$
\item incorporate the new data point $\mathbf{x}_{N + 1}$, which is now a simple rank-1 update on $\mathbf{A}$, so $\mathbf{A}^{-1}$ or $\chol(\mathbf{A})$ can again be updated efficiently in $\mathcal{O}(C^2)$ time
\end{enumerate}
We refer to Supplement~\ref{sec:SuppOnlineRidgeRegression} for more details.

With gradient descent trainable models, we maintain (an approximation to) the optimal weights $\mathbf{w}$ directly. When a new data point arrives, we expand the dimensionality of $\phi$ as described above. The previously optimal weights can be padded with $0$'s in any newly added dimensions, and then passed to the gradient descent method as initialization.

\section{LINK TO MONDRIAN FOREST}
\label{sec:ForestConnection}

We contrast Mondrian kernel with Mondrian forest \citep{MF, MFreg}, another non-linear learning method based on the Mondrian process. They both start by sampling $M$ independent Mondrians on $\Reals^D$ to provide $M$ independent partitions of the data. However, these partitions are then used differently in the two models:
\begin{itemize}
\item In a Mondrian forest, parameters of predictive distributions in each tree are fitted independently of all other trees. The prediction of the forest is the average prediction among the $M$ trees.
\item With Mondrian kernel, the weights of all random features are fitted jointly by a linear learning method.
\end{itemize}

Let $C^{(m)}$ count the leaves (non-empty partition cells) in the $m$-th Mondrian sample and let $C = \sum_{m = 1}^M C^{(m)}$ be the total number of leaves. Let $\bphi^{(m)}_n := \phi^{(m)}(\bx_n) \in \Reals^{C^{(m)}}$ be the indicator of the partition cell in the $m$-th sample into which the $n$-th data point falls (as in section~\ref{sec:MondrianKernel}). Also, as in equation~(\ref{eq:PhiConcat}), let $\bphi_n := \phi(\bx_n) \in \Reals^C$ be the normalized concatenated feature encoding of the $n$-th data point. Recall that each vector $\bphi_n \in \Reals^C$ contains exactly $M$ non-zero entries, all of which equal the normalizer $M^{-1/2}$.

For simplicity, we restrict our attention to ridge regression in this section and compare the learning objective functions of Mondrian kernel and Mondrian forest.

\subsection{MONDRIAN KERNEL OBJECTIVE}

The primal ridge regression problem in the feature space implied by $\phi$ is
\begin{equation*}
\min_{\bw \in \Reals^C}\;
\sum_{n = 1}^N (y_n - \bw^T \bphi_n)^2
+ \delta^2 \| \bw \|_2^2.
\end{equation*}
Decomposing $\bw = M^{-1/2} [ \bw^{(1)T} \cdots \bw^{(M)T}]^T$, so that each (rescaled) subvector $\bw^{(m)}$ corresponds to features from the $m$-th Mondrian, denoting by $\hat{y}_n^{(m)} := \bw^{(m)T} \bphi_n^{(m)}$ the ``contribution" of the $m$-th Mondrian to the prediction at the $n$-th data point, and writing $\text{loss}(y, \hat{y}) := (y - \hat{y})^2$, the Mondrian kernel objective function can be restated as
\begin{equation}
\min_{\bw \in \Reals^C}
\sum_{n = 1}^N \text{loss}\left(y_n, \frac{1}{M} \sum_{m = 1}^M \hat{y}_n^{(m)} \right)
+ \delta^2 \| \bw \|_2^2.
\label{eq:MondrianKernelObjective}
\end{equation}

\subsection{MONDRIAN FOREST OBJECTIVE}

Assuming a factorizing Gaussian prior over the leaves in each Mondrian tree (i.e., without the hierarchical smoothing used by \citet{MFreg}), the predictive mean parameters $\bw^{(m)}$ in the leaves of the $m$-th Mondrian tree are fitted by minimizing
\begin{equation*}
\min_{\bw^{(m)} \in \Reals^{C^{(m)}}}
\sum_{n = 1}^N (y_n - \bw^{(m)T} \bphi^{(m)}_n)^2
+ \gamma^2 \| \bw^{(m)} \|_2^2
\end{equation*}
where $\gamma^2$ is the ratio of noise and prior variance in the predictive model. The parameters $\bw^{(m)}$ are disjoint for different trees, so these $M$ independent optimization problems are equivalent to minimizing the average of the $M$ individual objectives. Writing $\hat{y}_n^{(m)} := \bw^{(m)T} \bphi_n^{(m)}$ for the $m$-th tree's prediction at the $n$-th data point and concatenating the parameters $\bw := M^{-1/2} [\bw^{(1)T} \cdots \bw^{(M)T}]^T$, the Mondrian forest objective can be stated as
\begin{equation}
\min_{\bw \in \Reals^C}
\sum_{n = 1}^N \frac{1}{M} \sum_{m = 1}^M \text{loss}(y_n, \hat{y}_n^{(m)})
+ \gamma^2 \| \bw \|_2^2.
\label{eq:MondrianForestObjective}
\end{equation}

\subsection{DISCUSSION}

Comparing \eqref{eq:MondrianKernelObjective} and \eqref{eq:MondrianForestObjective}, we see that subject to regularization parameters (priors) chosen compatibly, the two objectives only differ in the contribution of an individual data point $n$ to the total loss:
\begin{align*}
\text{Mondrian kernel:} \hspace{1em}
& \text{loss}\left(y_n, \frac{1}{M} \sum_{m = 1}^M \hat{y}_n^{(m)} \right)
\\
\text{Mondrian forest:} \hspace{1em}
& \frac{1}{M} \sum_{m = 1}^M \text{loss}(y_n, \hat{y}_n^{(m)})
\end{align*}
Specifically, the difference is in the order in which the averaging $\frac{1}{M} \sum_{m = 1}^M$ over Mondrian samples/trees and the non-linear $\text{loss}$ function are applied. In both models predictions are given by $\hat{y} = \frac{1}{M} \sum_{m = 1}^M \hat{y}^{(m)}$, so the Mondrian kernel objective is consistent with the aim of minimizing empirical loss on the training data, while the forest objective minimizes average loss across trees, not the loss of the actual prediction (when $M > 1$) \citep{ren_global_2015}.

\citet{ren_global_2015} address this inconsistency between learning and prediction by proposing to extend random forests with a \emph{global refinement} step that optimizes all tree parameters jointly, minimizing the empirical training loss. Our approximation of the Laplace kernel via the Mondrian kernel can be interpreted as implementing this joint parameter fitting step on top of Mondrian forest, revealing a new connection between random forests and kernel methods. 

\begin{figure}[H]
\centering
\includegraphics[width=0.95\columnwidth]{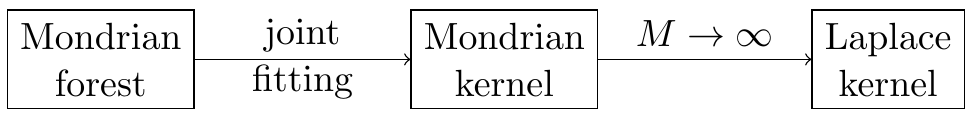}
\label{fig:ForestKernelConnection}
\end{figure}

\section{RELATED WORK}

The idea of \citet{rahimi2007random} to approximate shift-invariant kernels by constructing random features has been further developed by
\citet{fastfood} and \citet{yang_carte_2014}, providing a faster method of constructing the random features when the input dimension $D$ is high. The fast method of \citet{dai2014scalable} can adapt the number of random features, making it better-suited for streaming data. To the best of our knowledge, these methods require random features to be reconstructed from scratch for each new kernel width value; however, our solution allows us to efficiently learn this hyperparameter for the Laplace kernel. 

Decision forests are popular for black-box classification and regression thanks to their competitive accuracy and computational efficiency. The most popular variants are Breiman's Random Forest \citep{RF} and Extremely Randomized Trees \citep{ERT}.
\citet{infinitytheory} established a link between the Laplace kernel and random forests with an infinite number of trees, but unlike our work, made two additional strong assumptions, namely infinite data and a uniform distribution of features.
From a computational perspective, \citet{shen2006fast} approximated evaluation of an isotropic kernel using $kd$-trees, reducing computational complexity as well as memory requirements. \citet{davies2014random} constructed `supervised' kernels using random forests and demonstrated that this can lead to linear-time inference. We refer to \citep{scornet2015random}  for a recent discussion on the connection between decision forests and kernel methods. 

A key difference between decision forests and kernel methods is whether parameters are fit independently or jointly. In decision forests, the leaf node parameters for each tree are fit independently, whereas the weights of random features are fit jointly. \citet{scornet2015random} shows that random forests can be interpreted as adaptive kernel estimates and discusses the theoretical properties of fitting parameters jointly.  \citet{ren_global_2015} propose to extend random forests with a \emph{global refinement} step, optimizing all tree parameters jointly to minimize empirical training loss. 

The proposed Mondrian kernel establishes a link between Mondrian trees and Laplace kernel for finite data, without any assumptions on the distribution of the features. Unlike prior work, we exploit this connection to construct an adaptive random feature approximation and efficiently learn the kernel width. %

\section{EXPERIMENTS}

We conducted three sets of experiments, with these goals:
\begin{enumerate}
\item verify that Mondrian kernel approximates the Laplace kernel, and compare to other random feature generation schemes (Section~\ref{sec:ExperimentLaplaceKernelApproximation});
\item demonstrate usefulness of our efficient kernel width selection procedure, showing that it can quickly learn a suitable kernel width from data (Section~\ref{sec:FastKernelWidthLearning}); and
\item empirically compare the Mondrian kernel and Mondrian forests, supporting the insight into their relationship from Section~\ref{sec:ForestConnection} (Section~\ref{sec:ExperimentKernelVsForest}).
\end{enumerate}

With the exception of two experiments on synthetic data, we carried out our evaluation on the CPU  dataset from \citep{rahimi2007random}, containing $N = 6554$ training and $N_{\text{test}} = 819$ test points with $D = 21$ attributes. Note that the CPU dataset is an adversarial choice here, as \citet{rahimi2007random} report that random Fourier features perform better than binning schemes on this task. In all experiments, the ridge regularization constant was set to $\delta^2 = 10^{-4}$, the value used by \citet{rahimi2007random}, and the primal optimization problems were solved using stochastic gradient descent.

\subsection{LAPLACE KERNEL APPROXIMATION}
\label{sec:ExperimentLaplaceKernelApproximation}

\begin{figure}[t!]
\centering
\includegraphics[scale=0.5]{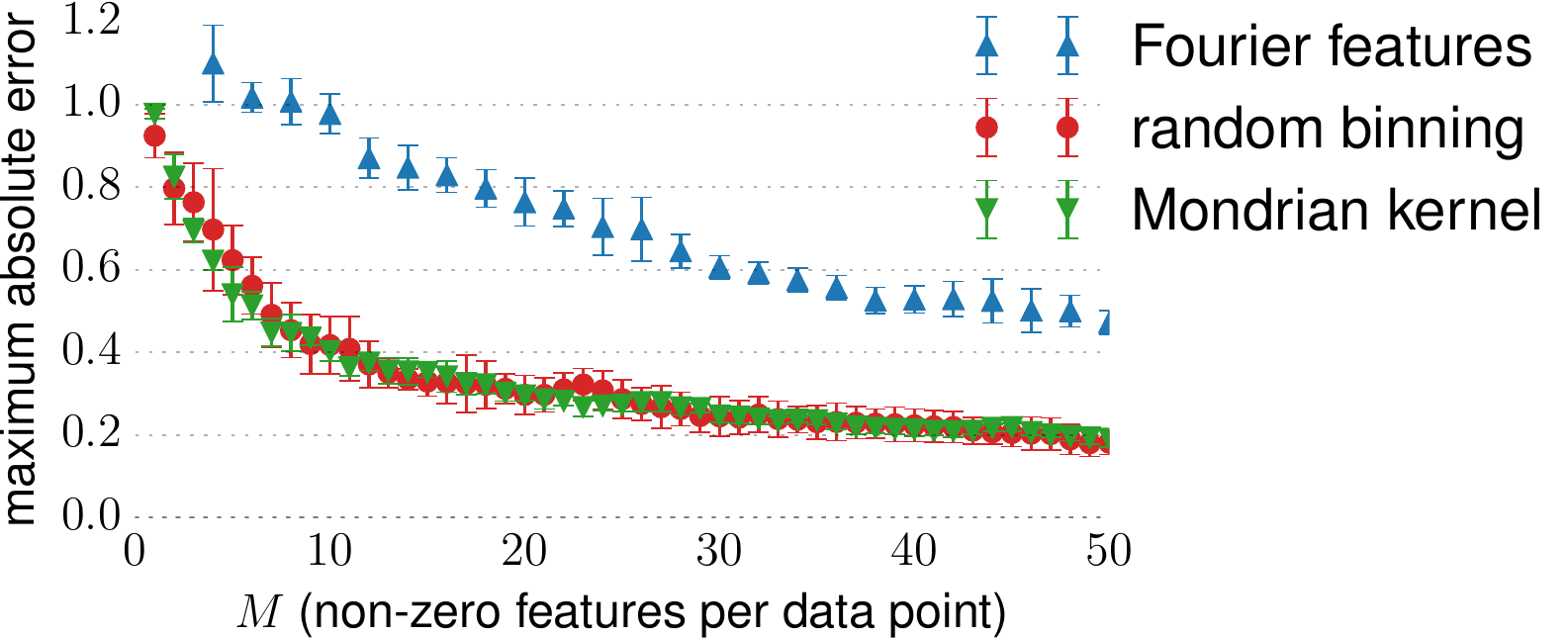}
\caption{Maximum absolute kernel approximation error on all pairs of $N = 100$ data points in $[0, 1]^2$.}
\label{fig:ExperimentLaplaceKernelApproximationDirect}
\end{figure}

First we examined the absolute kernel approximation error $|k_{\infty}(\cdot, \cdot) - k_M(\cdot, \cdot)|$ directly. To this end, we sampled $N = 100$ data points uniformly at random in the unit square $[0, 1]^2$ and computed the maximum absolute error over all $N^2$ pairs of points. The Laplace kernel $k_{\infty}$ and Mondrian kernels $k_M$ had a common lifetime (inverse width) $\lambda = 10$, so that several widths fit into the input domain $[0, 1]^2$. We repeated the experiment $5$ times for each value of $M$, showing the results in Figure~\ref{fig:ExperimentLaplaceKernelApproximationDirect}. We plot the maximum error against the number $M$ of non-zero features per data point, which is relevant for solvers such as Pegasos SVM \citep{shalev-shwartz_pegasos:_2007}, whose running time scales with the number of non-zero features per data point. Under this metric, the Mondrian kernel and Random binning converged to the Laplace kernel faster than random Fourier features, showing that in some cases they can be a useful option. (The error of Random Fourier features would decrease faster when measured against the \emph{total} number of features, as Mondrian kernel and Random binning generate sparse feature expansions.)

\begin{figure}[H]
\centering
\includegraphics[scale=0.5]{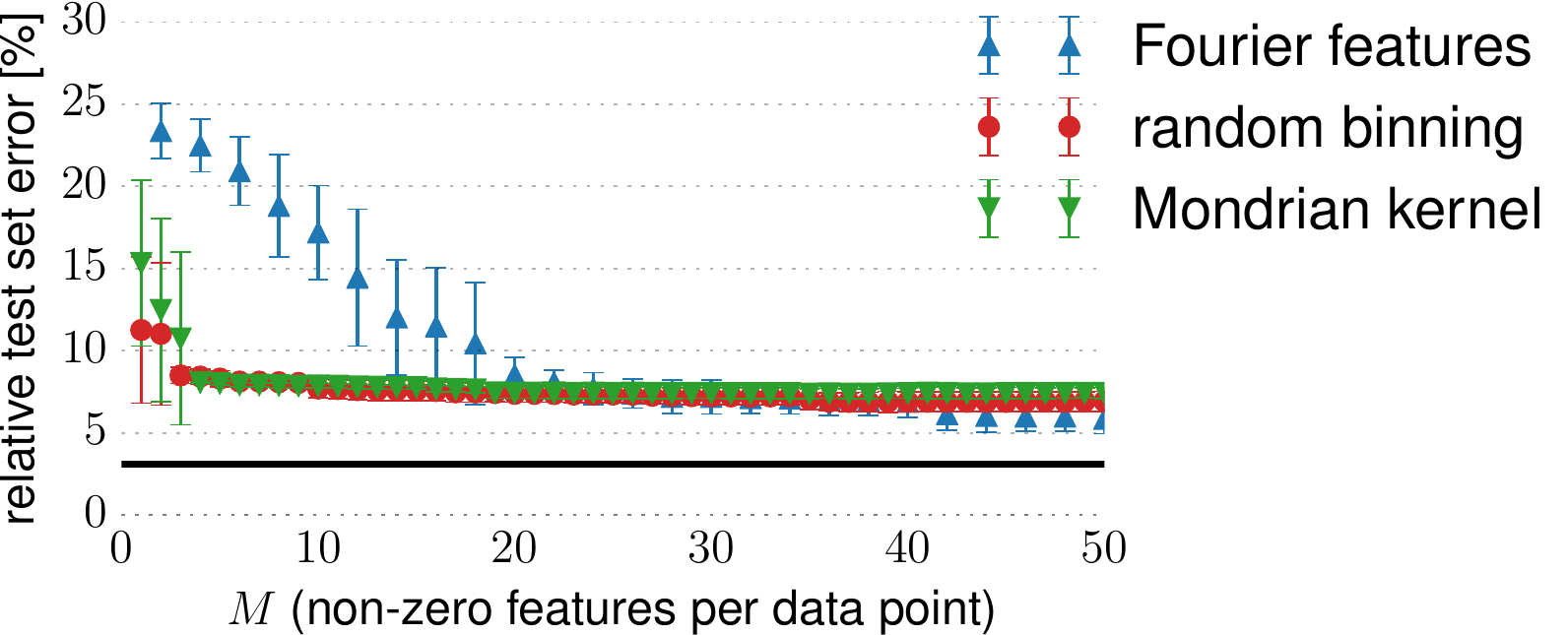}
\caption{Test set error on the CPU dataset. The horizontal line at $3.1\%$ indicates the error achieved with an exact, but expensive computation using the Laplace kernel.}
\label{fig:ExperimentLaplaceKernelApproximationCPU}
\end{figure}

Second, we examined the approximation error indirectly via test set error on the CPU dataset. We repeated the experiment $5$ times for each value of $M$ and show the results in Figure~\ref{fig:ExperimentLaplaceKernelApproximationCPU}. Even though Fourier features are better suited to this task, for a fast approximation with few ($M < 15$) non-zero features per data point, random binning and Mondrian kernel are still able to outperform the Fourier features.

\subsection{FAST KERNEL WIDTH LEARNING}
\label{sec:FastKernelWidthLearning}

\begin{figure}[b]
	\centering
	\includegraphics[scale=0.45]{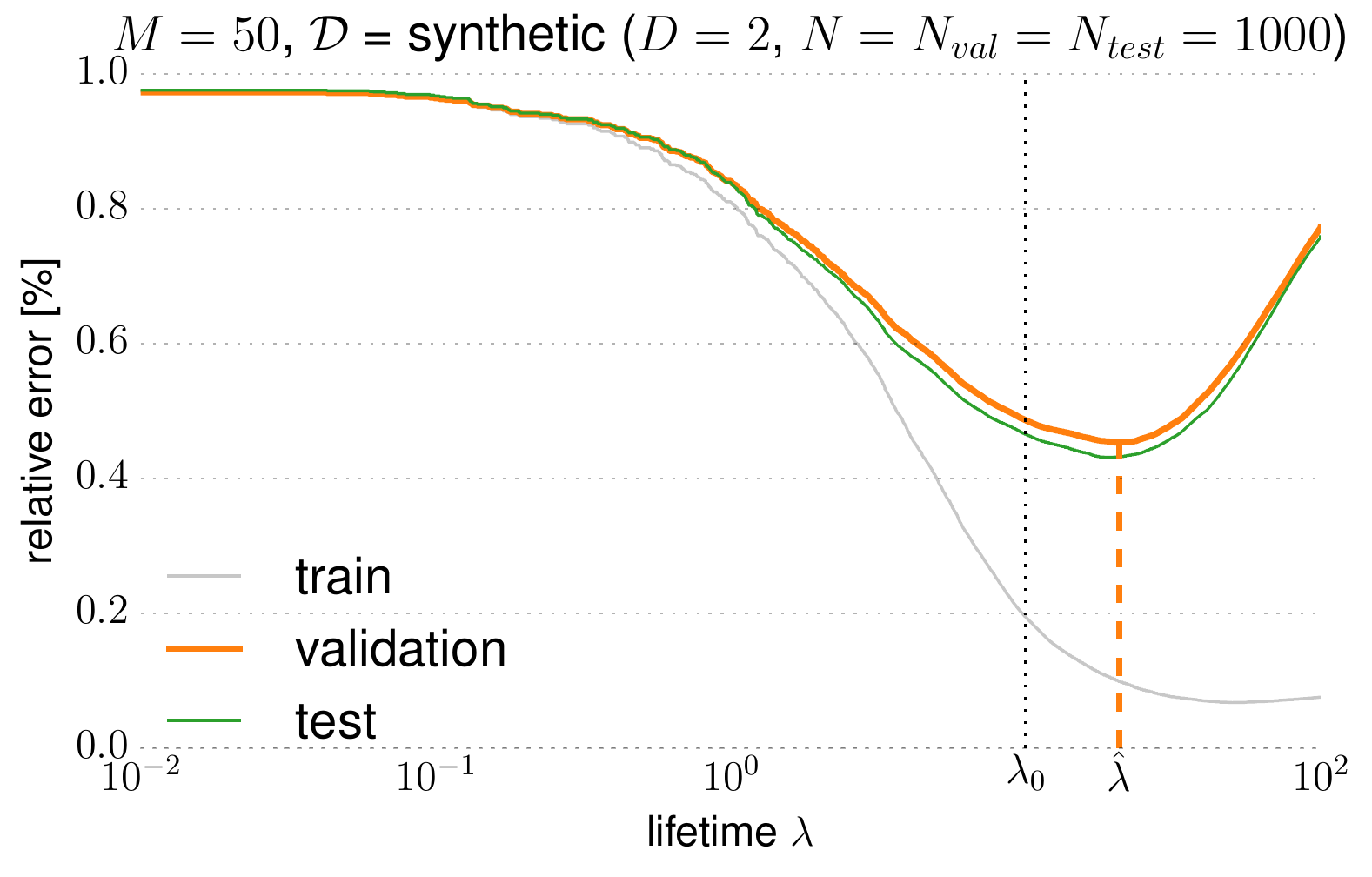}
	\caption{Recovering the ground truth lifetime $\lambda_0 = 10$ by selecting the value $\hat{\lambda} \approx 19$ minimizing validation set error.}
	\label{fig:ExperimentKernelWidthRecoverySynthetic}
\end{figure}

First, using a synthetic regression dataset generated from a Laplace kernel with known ground truth lifetime $\lambda_0 = 10$, we verified that the lifetime could be recovered using our kernel width selection procedure from Section~\ref{sec:LengthscaleLearning}. To this end, we let the procedure run until a terminal lifetime $\Lambda = 100$ and plotted the error on a held-out validation set as a function of the lifetime $\lambda$. The result in Figure~\ref{fig:ExperimentKernelWidthRecoverySynthetic} shows that the ground truth kernel lifetime $\lambda_0 = 10$ was recovered within an order of magnitude by selecting the lifetime $\hat{\lambda}$ minimizing validation set error. Moreover, this value of $\hat{\lambda}$ led to excellent performance on an independent test set.

\begin{figure}[t!]
	\centering
	\includegraphics[width=\linewidth]{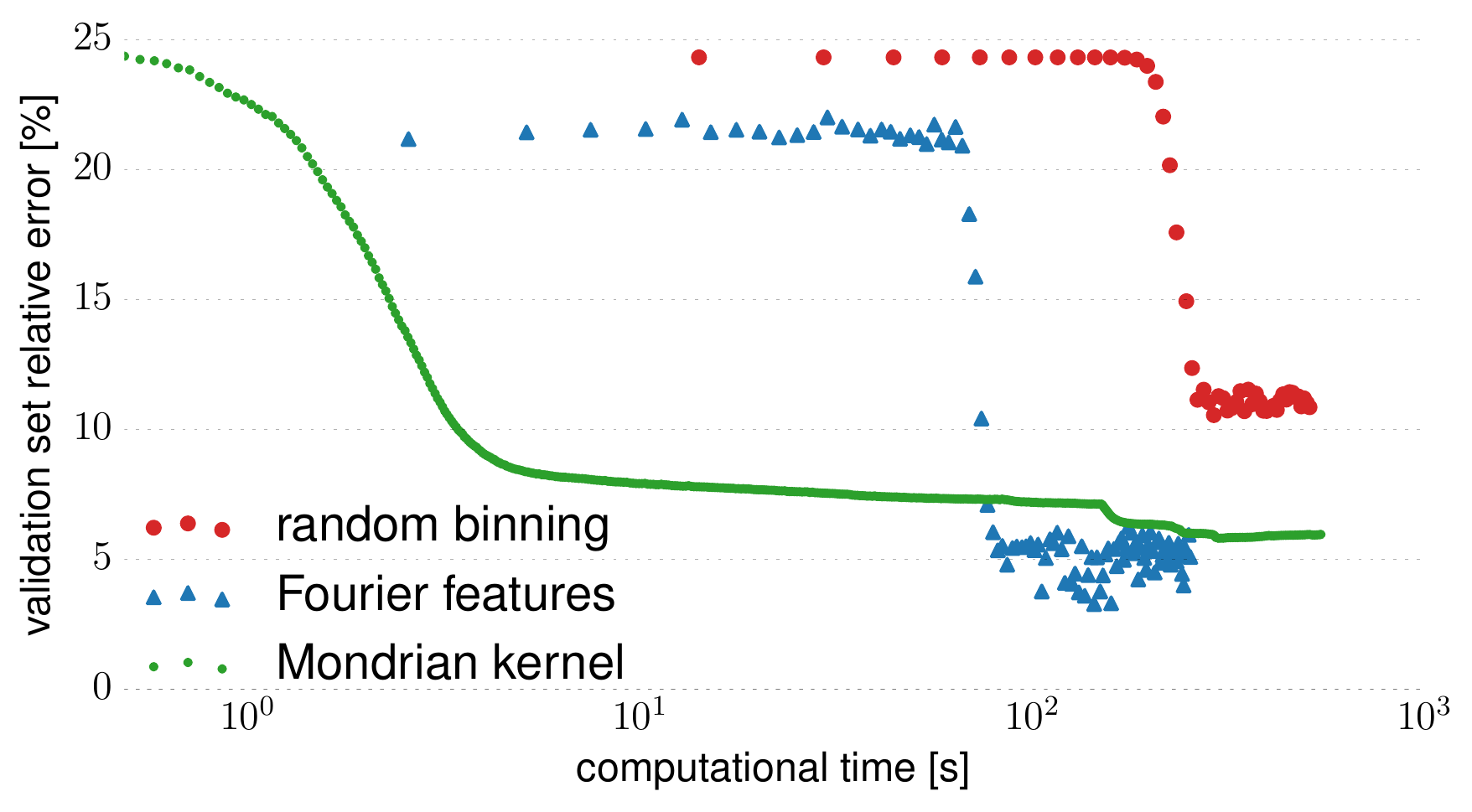}
	\caption{Validation set error as a function of computation time. Even though Fourier features are better suited to the CPU dataset \citep{rahimi2007random} and eventually outperform the Mondrian kernel, the latter discovers suitable kernel widths at least an order of magnitude faster.}
	\label{fig:ExperimentTimeToBestSolution}
\end{figure}

\pagebreak
Second, we evaluated our kernel width selection procedure on the CPU dataset in order to demonstrate its practical usefulness. While the Mondrian kernel allows to efficiently sweep through lifetimes $\lambda$, Fourier features and random binning need to be reconstructed and retrained for each attempted lifetime value. We started the Fourier features and random binning at $\lambda = 1$, and in each step, we either doubled the maximum lifetime or halved the minimum lifetime considered so far, based on which direction seemed more promising. Once a good performing lifetime was found, we further optimized using a binary search procedure. All schemes were set to generate $M = 350$ non-zero features per datapoint. Figure~\ref{fig:ExperimentTimeToBestSolution} shows the performance of each scheme on a held-out validation set as a function of computation time. The result suggests that our kernel width learning procedure can be used to discover suitable lifetimes (inverse kernel widths) at least an order of magnitude faster than random Fourier features or random binning.

\subsection{MONDRIAN KERNEL VS FOREST}
\label{sec:ExperimentKernelVsForest}

\begin{figure}[t!]
  \centering
  \includegraphics[scale=0.464]{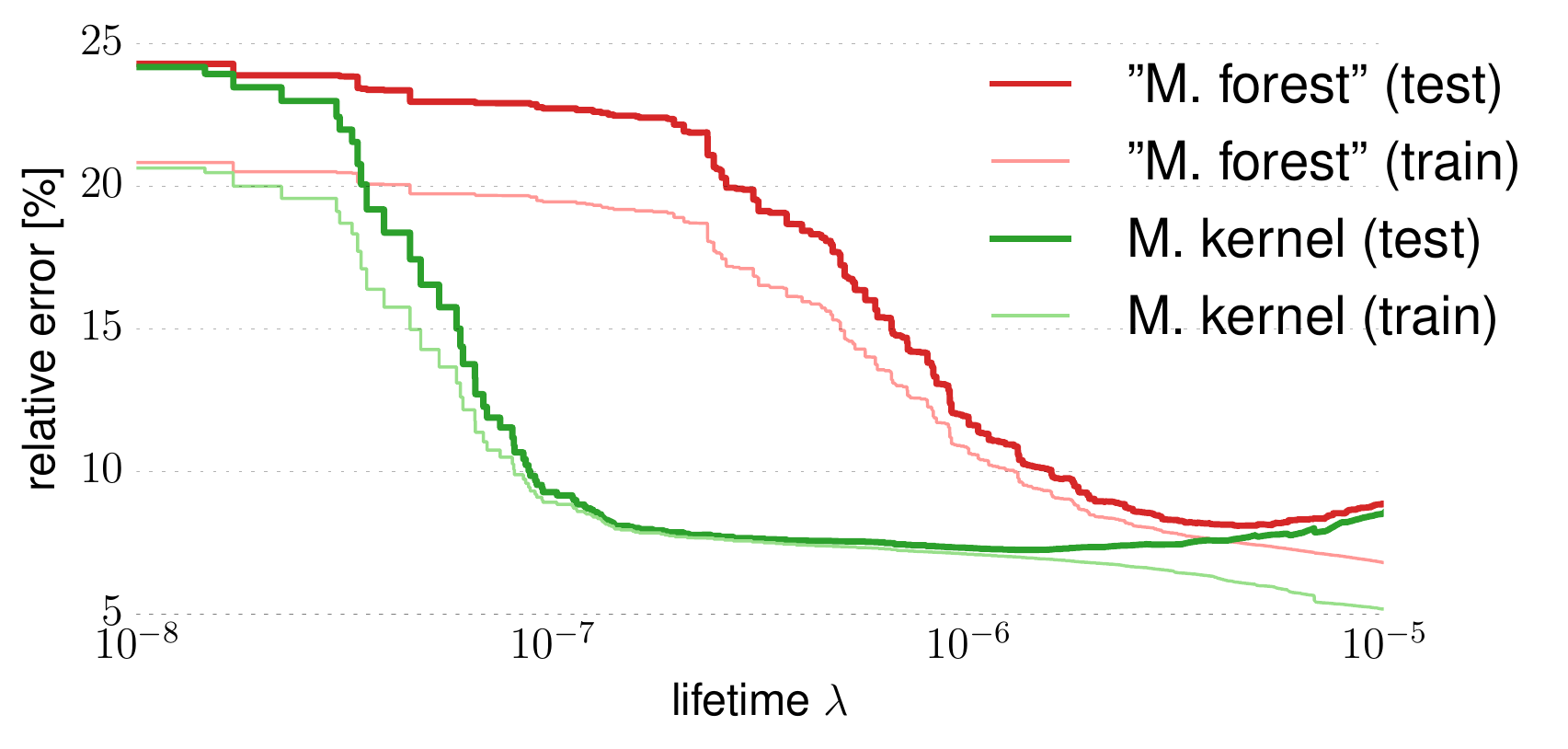}
  \caption{Comparison of Mondrian kernel and Mondrian forest models based on the same set of Mondrian samples.}
  \label{fig:ExperimentMondrianKernelVsForest}
\end{figure}

\begin{figure}[t!]
  \centering
  \includegraphics[width=\linewidth]{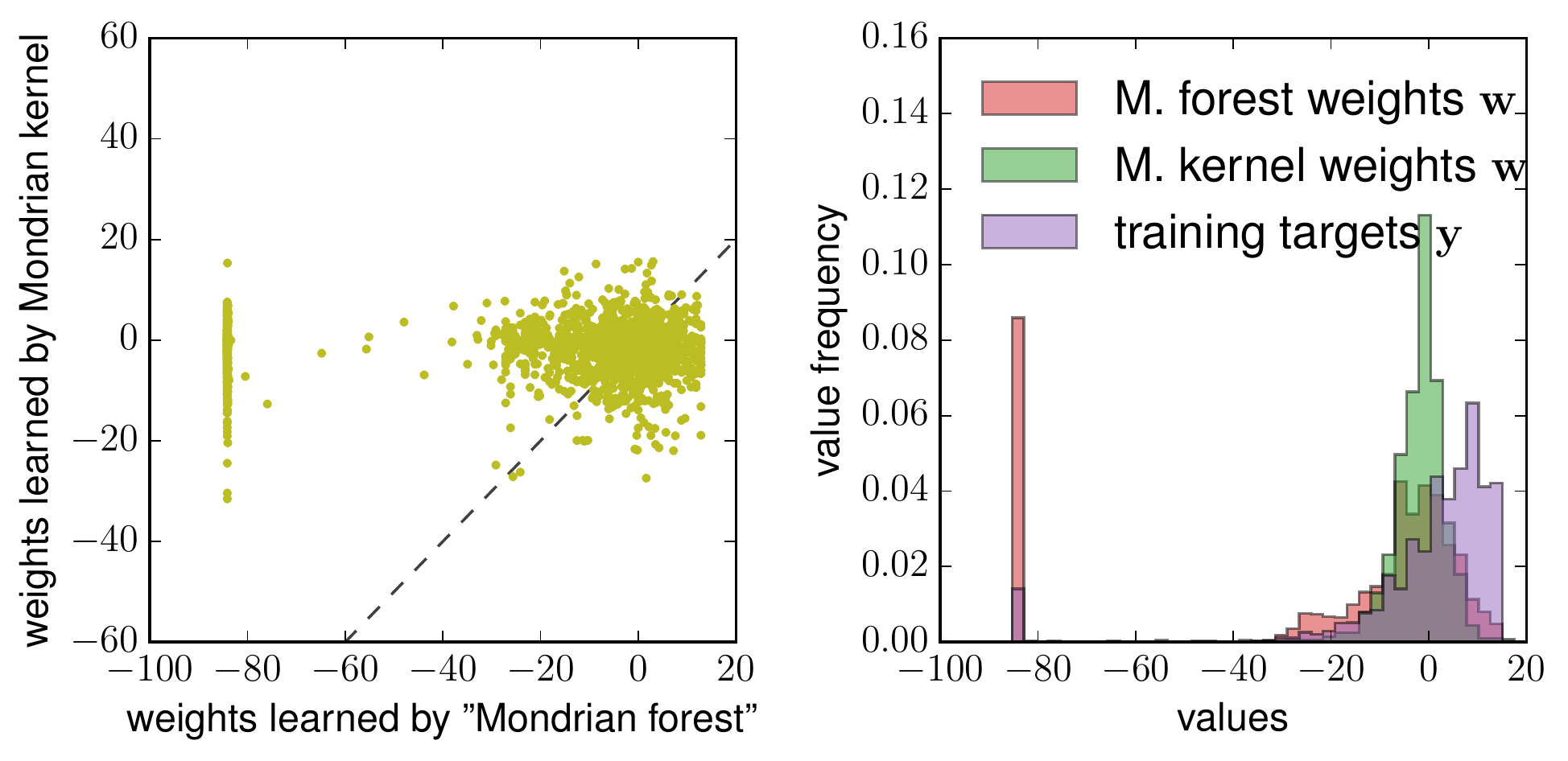}
  \caption{Weights learned by Mondrian forest and Mondrian kernel at the lifetime $\lambda = 2 \times 10^{-6}$ in Figure~\ref{fig:ExperimentMondrianKernelVsForest}.}
  \label{fig:ExperimentMondrianKernelVsForestWeights}
\end{figure}

We compared the performance of Mondrian kernel and ``Mondrian forest" (quotes due to omission of hierarchical smoothing) based on the same $M = 50$ Mondrian samples, using the CPU dataset and varying the lifetime $\lambda$.
Recall that higher values of $\lambda$ lead to more refined Mondrian partitions, allowing more structure in the data to be modeled, but also increasing the risk of overfitting.
Figure~\ref{fig:ExperimentMondrianKernelVsForest} shows that Mondrian kernel exploits the joint fitting of parameters corresponding to different trees and achieves a lower test error at lower lifetime values, thus producing a more compact solution based on simpler partitions. Figure~\ref{fig:ExperimentMondrianKernelVsForestWeights} shows the parameter values learned by Mondrian kernel and Mondrian forest at the lifetime $\lambda = 2 \times 10^{-6}$. 
The distribution of weights learned by Mondrian kernel is more peaked around $0$, as the joint fiting allows achieving more extreme predictions by adding together several smaller weights.

\section{CONCLUSION}

We presented the Mondrian kernel, a fast approximation to the Laplace kernel that admits efficient kernel width selection.
When a different kernel or a different approximation is used, our procedure can provide a fast and simple way of initializing the kernel width for further optimization.
While a Gaussian kernel is often considered a default choice, in many situations it imposes an inappropriately strong smoothness assumption on the modelled function and the Laplace kernel may in fact be a preferable option.

Our approach revealed a novel link between the Mondrian process and the Laplace kernel. We leave the discovery of similar links involving other kernels for future work.

\subsubsection*{Acknowledgements}

We would like to thank Nilesh Tripuraneni for useful discussions.
Part of this research was carried out while MB was at the University of Oxford.
BL gratefully acknowledges generous funding from the Gatsby Charitable Foundation.
ZG acknowledges funding from the Alan Turing Institute, Google, Microsoft Research and EPSRC Grant EP/N014162/1.
DMR is supported by an NSERC Discovery Grant.
YWT's research leading to these results has received funding from the European Research Council under the European Union's Seventh Framework Programme (FP7/2007-2013) ERC grant agreement no. 617071.

\subsubsection*{References}
\bibliography{mondrian_kernel_UAI}

\begin{thebibliography}{20}
\providecommand{\natexlab}[1]{#1}
\providecommand{\url}[1]{\texttt{#1}}
\expandafter\ifx\csname urlstyle\endcsname\relax
  \providecommand{\doi}[1]{doi: #1}\else
  \providecommand{\doi}{doi: \begingroup \urlstyle{rm}\Url}\fi

\bibitem[Breiman(2000)]{infinitytheory}
L.~Breiman.
\newblock Some infinity theory for predictor ensembles.
\newblock Technical report, University of California at Berkeley, 2000.

\bibitem[Breiman(2001)]{RF}
L.~Breiman.
\newblock Random forests.
\newblock \emph{Mach. Learn.}, 45:\penalty0 5--32, 2001.

\bibitem[Criminisi et~al.(2012)Criminisi, Shotton, and Konukoglu]{DF}
A.~Criminisi, J.~Shotton, and E.~Konukoglu.
\newblock Decision forests: A unified framework for classification, regression,
  density estimation, manifold learning and semi-supervised learning.
\newblock \emph{Found. Trends Comput. Graphics and Vision}, 2012.

\bibitem[Dai et~al.(2014)Dai, Xie, He, Liang, Raj, Balcan, and
  Song]{dai2014scalable}
B.~Dai, B.~Xie, N.~He, Y.~Liang, A.~Raj, M.-F.~F. Balcan, and L.~Song.
\newblock Scalable kernel methods via doubly stochastic gradients.
\newblock In \emph{Adv. Neural Information Proc. Systems (NIPS)}, 2014.

\bibitem[Davies and Ghahramani(2014)]{davies2014random}
A.~Davies and Z.~Ghahramani.
\newblock The random forest kernel and other kernels for big data from random
  partitions.
\newblock \emph{arXiv preprint arXiv:1402.4293v1}, 2014.

\bibitem[Geurts et~al.(2006)Geurts, Ernst, and Wehenkel]{ERT}
P.~Geurts, D.~Ernst, and L.~Wehenkel.
\newblock Extremely randomized trees.
\newblock \emph{Mach. Learn.}, 63\penalty0 (1):\penalty0 3--42, 2006.

\bibitem[Lakshminarayanan et~al.(2014)Lakshminarayanan, Roy, and Teh]{MF}
B.~Lakshminarayanan, D.~M. Roy, and Y.~W. Teh.
\newblock Mondrian forests: Efficient online random forests.
\newblock In \emph{Adv. Neural Information Proc. Systems (NIPS)}, 2014.

\bibitem[Lakshminarayanan et~al.(2016)Lakshminarayanan, Roy, and Teh]{MFreg}
B.~Lakshminarayanan, D.~M. Roy, and Y.~W. Teh.
\newblock Mondrian forests for large scale regression when uncertainty matters.
\newblock In \emph{Int. Conf. Artificial Intelligence Stat. (AISTATS)}, 2016.

\bibitem[Le et~al.(2013)Le, Sarl{\'o}s, and Smola]{fastfood}
Q.~Le, T.~Sarl{\'o}s, and A.~Smola.
\newblock Fastfood-approximating kernel expansions in loglinear time.
\newblock In \emph{Proc. Int. Conf. Mach. Learn. (ICML)}, 2013.

\bibitem[Rahimi and Recht(2007)]{rahimi2007random}
A.~Rahimi and B.~Recht.
\newblock Random features for large-scale kernel machines.
\newblock In \emph{Adv. Neural Information Proc. Systems (NIPS)}, 2007.

\bibitem[Ren et~al.(2015)Ren, Cao, Wei, and Sun]{ren_global_2015}
S.~Ren, X.~Cao, Y.~Wei, and J.~Sun.
\newblock Global refinement of random forest.
\newblock In \emph{Proc. {IEEE} {Conference} on {Computer} {Vision} and
  {Pattern} {Recognition}}, pages 723--730, 2015.

\bibitem[Roy(2011)]{RoyThesis}
D.~M. Roy.
\newblock \emph{Computability, inference and modeling in probabilistic
  programming}.
\newblock PhD thesis, Massachusetts Institute of Technology, 2011.

\bibitem[Roy and Teh(2009)]{RT09}
D.~M. Roy and Y.~W. Teh.
\newblock {The {M}ondrian process}.
\newblock In \emph{Adv. Neural Information Proc. Systems (NIPS)}, 2009.

\bibitem[Sch\"olkopf and Smola(2001)]{scholkopf_learning_2001}
B.~Sch\"olkopf and A.~J. Smola.
\newblock \emph{Learning with {Kernels}: {Support} {Vector} {Machines},
  {Regularization}, {Optimization}, and {Beyond}}.
\newblock MIT Press, Cambridge, MA, USA, 2001.
\newblock ISBN 978-0-262-19475-4.

\bibitem[Scornet(2015)]{scornet2015random}
E.~Scornet.
\newblock Random forests and kernel methods.
\newblock \emph{arXiv preprint arXiv:1502.03836v2}, 2015.

\bibitem[Seeger(2003)]{seeger_bayesian_2003}
M.~Seeger.
\newblock \emph{Bayesian {Gaussian} {Process} {Models}: {PAC}-{Bayesian}
  {Generalisation} {Error} {Bounds} and {Sparse} {Approximations}}.
\newblock PhD thesis, University of Edinburgh, 2003.

\bibitem[Seeger(2004)]{EPFL-REPORT-161468}
M.~Seeger.
\newblock Low rank updates for the {C}holesky decomposition.
\newblock Technical report, University of California at Berkeley, 2004.

\bibitem[Shalev-Shwartz et~al.(2007)Shalev-Shwartz, Singer, and
  Srebro]{shalev-shwartz_pegasos:_2007}
S.~Shalev-Shwartz, Y.~Singer, and N.~Srebro.
\newblock Pegasos: {Primal} {Estimated} sub-{GrAdient} {SOlver} for {SVM}.
\newblock In \emph{Proc. Int. Conf. Mach. Learn. (ICML)}, 2007.

\bibitem[Shen et~al.(2006)Shen, Ng, and Seeger]{shen2006fast}
Y.~Shen, A.~Ng, and M.~Seeger.
\newblock Fast {G}aussian process regression using {KD}-trees.
\newblock In \emph{Adv. Neural Information Proc. Systems (NIPS)}, 2006.

\bibitem[Yang et~al.(2015)Yang, Smola, Song, and Wilson]{yang_carte_2014}
Z.~Yang, A.~J. Smola, L.~Song, and A.~G. Wilson.
\newblock A la {Carte} - {Learning} {Fast} {Kernels}.
\newblock In \emph{Int. Conf. Artificial Intelligence Stat. (AISTATS)}, 2015.

\end{thebibliography}

\newpage
{\centerline{\Large{\textbf{The Mondrian Kernel}}}}
{\centerline{\Large{\textbf{Supplementary material}}}}
\appendix

\section{Proofs}
\label{sec:AppendixConvergenceBounds}

\begin{definition} The \emph{linear dimension} of an axis-aligned box $\calX = \calX_1 \times \cdots \times \calX_D \subseteq \Reals^D$ is $|\calX| := |\calX_1| + \cdots + |\calX_D|$.
\end{definition}

Our first result is a tail bound on the number of partition cells generated by a Mondrian process. We will use it as a Lemma in Proposition~\ref{prop:MondrianKernelUniformConvergenceWithProof}, but it also confirms that with probability 1, the Mondrian process does not explode (does not generate infinitely many partition cells in finite time).

\begin{proposition}
\label{prop:MondrianPartitionsTailBound}
Let $\mathcal{M}$ be a Mondrian process on an axis-aligned box $\calX$. For $t \geq 0$, let $N_t$ be the number of partition cells generated by $\mathcal{M}$ until time $t$. Then
\begin{equation*}
\forall{n \in \Reals_{+}}
\hspace{2em}
\IP[N_t > n]
\leq \frac{e^{|\calX| t}}{n}.
\end{equation*}
In particular, the Mondrian process does not explode.
\begin{proof}
At any time $s$, by lack of memory of the exponential distribution, the residual time until a partition cell $c$ splits into two has $\Exp(|c|)$ distribution and is independent of all other cells by construction of the Mondrian process. As $|c| \leq |\calX|$, this cell splitting process is dominated by a Yule process with birth rate $|\calX|$. The number $\tilde{N}_t$ of individuals at time $t$ of a Yule process with birth rate $|\calX|$ has geometric distribution with mean $e^{|\calX| t}$ and Markov's inequality yields
\begin{equation*}
\IP[N_t > n]
\leq \IP[\tilde{N}_t > n]
\leq \frac{e^{|\calX| t}}{n}.
\end{equation*}
as claimed. Hence $\IP[ N_t = \infty ] = \lim_{n \to \infty} \IP[ N_t > n ] = 0$ for any $t$.
\end{proof}
\end{proposition}

We define an $\varepsilon$-grid covering a (closed) interval as a set of points at most $\varepsilon$ distance apart, including the boundary points, and with minimal possible cardinality:

\begin{definition}
Let $\calX_1 = [a_1, b_1]$ be an interval of length $|\calX_1| = b_1 - a_1$ and let $0 < \varepsilon < |\calX_1|$. Define $K := \lceil \frac{|\calX_1|}{\varepsilon} \rceil$. An \emph{$\varepsilon$-grid} covering $\calX_1$ is a set $\mathcal{U}_1$ of $K + 1$ points $u_0 < u_1 < \cdots < u_K$ in $\calX_1$ such that $u_0 = a_1$, $u_K = b_1$ and $|u_i - u_{i - 1}| \leq \varepsilon$ for all $1 \leq i \leq K$.
\label{def:Grid1D}
\end{definition}

Note that such an $\varepsilon$-grid exists by our choice of $K$, as we can take, e.g., $u_i = i \varepsilon$ for $1 \leq i < K$. The next lemma bounds the probability that two arrivals of a Poisson process running on a bounded interval occur between two consecutive points of an $\varepsilon$-grid covering that interval.

\begin{lemma}
\label{lem:PoissonMeshing}
Consider a Poisson process with rate $\lambda$ running on a bounded interval $[0, L]$. Let $\mathcal{U}$ be an $\varepsilon$-grid covering of $[0, L]$. Then the probability that two or more arrivals of the process occur between any two consecutive points of $\mathcal{U}$ is at most $2 \lambda^2 L \varepsilon$.
\begin{proof}
As the distance between any two consecutive points of the $\varepsilon$-grid is at most $\varepsilon$ by definition, the number of arrivals in a line segment between such two points is dominated by a Poisson random variable with mean $\lambda \varepsilon$. As there are $\lceil \frac{L}{\varepsilon} \rceil$ such segments, the sought probability $p$ can be upper bounded using a union bound as 
\begin{equation*}
p \leq \left\lceil \frac{L}{\varepsilon} \right\rceil
     \left( 1 - e^{- \lambda \varepsilon} - e^{- \lambda \varepsilon} \lambda \varepsilon \right)
\end{equation*}
and using $1 - e^{-x} \leq x$ twice, we obtain as claimed
\begin{equation*}
p \leq
  \left\lceil \frac{L}{\varepsilon} \right\rceil
     \left( \lambda \varepsilon - e^{- \lambda \varepsilon} \lambda \varepsilon \right)
  \leq
  \left\lceil \frac{L}{\varepsilon} \right\rceil
     \left( \lambda \varepsilon \right)^2
  \leq
  2 L \lambda^2 \varepsilon.
\qedhere
\end{equation*}
\end{proof}
\end{lemma}

Definition~\ref{def:Grid1D} also set us up for defining the concept of an $\varepsilon$-grid on higher-dimensional axis-aligned boxes:

\begin{definition}
Let $\calX = \calX_1 \times \cdots \times \calX_D \subseteq \Reals^D$ be an axis-aligned box and let $\varepsilon > 0$. An \emph{$\varepsilon$-grid} covering $\calX$ is a cartesian product $\mathcal{U} = \mathcal{U}_1 \times \cdots \times \mathcal{U}_D$, where each $\mathcal{U}_d$ is an $\varepsilon$-grid covering of $\calX_d$ in the sense of Definition~\ref{def:Grid1D}.
\label{def:GridD}
\end{definition}

\begin{proposition} For any bounded input domain $\calX \subseteq \Reals^D$ and $\delta > 0$, as $M \to \infty$,
\begin{align*}
&\;\;\;\;\;
  \IP\left[ \sup_{\bx, \bx' \in \calX} \left| k_M(\bx, \bx') - k_{\infty}(\bx, \bx') \right| > \delta \right]
\\ &=
  \mathcal{O}\left( M^{2/3} e^{- M \delta^2 / (12D + 2)} \right).
\end{align*}
\begin{proof} By extending $\calX$ if necessary, we may assume without loss of generality that $\calX$ is an axis-aligned box with linear dimension $|\calX|$.

Recall that a Mondrian kernel of order $M$ corresponds to a random features obtained from $M$ independent Mondrians with lifetime $\lambda$. Let $\mathcal{U}$ be an $\varepsilon$-grid covering $\calX$, where $\varepsilon > 0$ will be specified later. The proof will upper bound the probability of the following three ``bad" events:
\begin{itemize}[leftmargin=*,label={}]
\item $A_1$
:= \{ {any of the $M$ Mondrian samples contains more than $n$ partition cells} \}
\item $A_2$
:= \{ {the common refinement of the $M$ Mondrian partitions, disregarding any potential cuts after generating $n$ cells in one Mondrian, has a partition cell that does not contain an element of $\mathcal{U}$} \}
\item
$A_3$
:= \{ {$\frac{\delta}{2}$-approximation fails on $\mathcal{U}$, i.e., for some $\mathbf{u}_1$, $\mathbf{u}_2 \in \mathcal{U}$, $|k_M(\mathbf{u}_1, \mathbf{u}_2) - k_{\infty}(\mathbf{u}_1, \mathbf{u}_2)| > \frac{\delta}{2}$ } \}
\end{itemize}
The constant $n \in \Reals{+}$ will be specified (optimized) later. Note that $A_1 \cap A_2$ implies that all partition cells in the common refinement of all $M$ Mondrian partitions contain a grid point from $\mathcal{U}$. Since $k_M$ is constant in each such cell, making $\varepsilon$ small enough, smoothness of the Laplace kernel $k_{\infty}$ will ensure that if $A_3^c$ holds then $\delta$-approximation holds throughout $\calX$.

Proposition~\ref{prop:MondrianPartitionsTailBound} and a union bound over the $M$ Mondrian samples give immediately that
\begin{equation*}
\IP(A_1)
\leq M \frac{e^{|\mathcal{X}| \lambda}}{n}.
\end{equation*}

Note that the $\varepsilon$-grid $\mathcal{U}$ contains at most $( 2 | \calX | / \varepsilon )^D$ grid points. Hoeffding's inequality and a union bound over all pairs of grid points gives for any $\varepsilon > 0$:
\begin{equation*}
\IP(A_3)
\leq \left[ \left( 2 \frac{| \calX |}{\varepsilon} \right)^D \right]^2 \left[ 2 \exp\left(- M \delta^2 / 2 \right) \right].
\end{equation*}

To upper bound the probability of $A_2$, note that at any time $t < \lambda$, in each partition cell generated so far by any of the $M$ Mondrian processes, an exponential clock is associated to each dimension $d$ of the cell, and if that clock rings, the cell is split at a random location $a$ by a hyperplane lying in dimension $d$. Consider the point process obtained by projecting the cut points from all partition cells onto their respective coordinate axes. If each Mondrian process generates no more than than $n$ partition cells until its lifetime $\lambda$ is exhausted, the cut points on the $d$-th coordinate axis come from at most $M n$ partition cells, each having width at most $|\calX_d|$ in dimension $d$. Therefore this point process on the $d$-th coordinate axis can be thought of as taking a suitable subset of points generated by a Poisson point process with intensity $M n |\calX_d| \lambda$. Thus by Lemma~\ref{lem:PoissonMeshing}, the probability that two cut points in dimension $d$ fall between two adjacent coordinates of the $\varepsilon$-grid $\mathcal{U}$ is upper bounded by $2 (M n \lambda)^2 |\calX_d| \varepsilon$. Observe that if this does not happen in any of the $D$ dimensions then all partition cells in the common refinement must contain a grid point from $\mathcal{U}$. Hence, taking the union bound over all $D$ dimensions,
\begin{equation*}
\IP(A_2)
\leq \sum_{d = 1}^D 2 (M n \lambda)^2 |\calX_d| \varepsilon
= 2 (M n \lambda)^2 |\calX| \varepsilon.
\end{equation*}

Thus the probability of a ``bad" event occuring is at most
\begin{align*}
& \IP(A_1 \cup A_2 \cup A_3)
\\ & \leq
  \IP(A_1) + \IP(A_2) + \IP(A_3)
\\ & \leq
  M \frac{e^{|\mathcal{X}| \lambda}}{n}
+ 2 (M n \lambda)^2 |\calX| \varepsilon
  + 2 \left(2 \frac{| \calX |}{\varepsilon} \right)^{2D} e^{- M \delta^2 / 2}.
\end{align*}
and minimizing over $n \in \Reals_{+}$ gives
\begin{align*}
&
  \IP(A_1 \cup A_2 \cup A_3)
\\ &\leq
    \left( 4 \lambda^2 M^2 |\calX| \varepsilon e^{2 \lambda |\calX|} \right)^{1/3}
  + 2 \left( \frac{|\calX|}{\varepsilon} \right)^{2D} e^{- M \delta^2 / 2}.
\end{align*}
If $A_1 \cap A_2$ holds then each cell in the common refinement of the $M$ Mondrian partitions contains an element of the $\varepsilon$-grid $\mathcal{U}$, and the Laplace kernel of lifetime $\lambda$ changes by at most $1 - e^{-D \lambda \varepsilon}$ when moving from any point in $\calX$ to the nearest grid point in its partition cell (in the common refinement). Therefore, as long as $2(1 - e^{-D \lambda \varepsilon}) < \frac{\delta}{2}$ (i.e., $\varepsilon \leq \frac{1}{\lambda D} \ln(1 - \frac{\delta}{4})$), the event $(A_1 \cup A_2 \cup A_3)^c$ implies that $\delta$-approximation holds throughout $\calX$. The upper bound on $\IP(A_1 \cup A_2 \cup A_3)$ above is minimized for
\begin{equation*}
\varepsilon_0
= \left( \frac{12 D |\calX|^{2D} e^{-M\delta^2/2}}{(4 \lambda |\calX|)^{1/3} e^{2 \lambda |\calX| / 3}} \right)
\end{equation*}
which tends to $0$ as $M \to \infty$ and so for large enough $M$, we do have $\varepsilon_0 \leq \frac{1}{\lambda D} \ln(1 - \frac{\delta}{4})$. For these large enough $M$ it then holds that
\begin{align*}
&%
  \IP\left[ \sup_{\bx, \bx' \in \calX} \left|\phi(\bx)^T \phi(\bx') - k(\bx, \bx') \right| > \delta \right]
\\ &\leq 
  \IP(A_1 \cup A_2 \cup A_3)
\\ &\leq
    \left( 4 \lambda^2 M^2 |\calX| \varepsilon_0 e^{2 \lambda |\calX|} \right)^{1/3}
  + 2 \left( \frac{|\calX|}{\varepsilon_0} \right)^{2D} e^{- M \delta^2 / 2}
\\ &=
  \left( 2^{1/(2D)} 4 \lambda^2 M^2 |\calX|^2 e^{2 \lambda L} / D \right)^{1/(3+1/2D)}
  e^{- \frac{M \delta^2}{12 D + 2}}
\\ &\in
 \mathcal{O}\left( M^{2/3} e^{- \frac{M \delta^2}{12 D + 2}} \right).
 \qedhere
\end{align*}
\end{proof}
\label{prop:MondrianKernelUniformConvergenceWithProof}
\end{proposition}

\begin{proposition} In a Mondrian regression forest with a factorizing Gaussian prior over leaf predictions, the learning objective function can be stated as
\begin{equation*}
\min_{\bw \in \Reals^C}
\sum_{n = 1}^N \frac{1}{M} \sum_{m = 1}^M \text{loss}(y_n, \hat{y}_n^{(m)})
+ \gamma^2 \| \bw \|_2^2.
\end{equation*}
\begin{proof}
The predictive mean parameters $\bw^{(m)}$ in the leaves of the $m$-th tree are fitted by solving
\begin{equation*}
\min_{\bw^{(m)} \in \Reals^{C^{(m)}}}
\sum_{n = 1}^N (y_n - \bw^{(m)T} \bphi^{(m)}_n)^2
+ \gamma^2 \| \bw^{(m)} \|_2^2
\end{equation*}
where $\gamma^2$ is the ratio of noise and prior variance in the predictive model. The parameters $\bw^{(m)}$ are disjoint for different trees, so these $M$ independent optimization problems are equivalent to minimizing the average
\begin{equation*}
\min_{\bw^{(1)}, \ldots, \bw^{(M)}}
\frac{1}{M} \sum_{m = 1}^M \left( \sum_{n = 1}^N (y_n - \hat{y}_n^{(m)})^2
+ \gamma^2 \| \bw^{(m)} \|_2^2 \right)
\end{equation*}
where $\hat{y}_n^{(m)} := \bw^{(m)T} \bphi_n^{(m)}$ is the $m$-th tree's prediction at data point $n$. Rewriting in terms of the squared loss $\text{loss}(y, \hat{y}) := (y - \hat{y})^2$ and the normalized concatenated weights $\bw := M^{-1/2} [\bw^{(1)T} \cdots \bw^{(M)T}]^T$, the learning objective function becomes
\begin{equation*}
\min_{\bw \in \Reals^C}
\sum_{n = 1}^N \frac{1}{M} \sum_{m = 1}^M \text{loss}(y_n, \hat{y}_n^{(m)})
+ \gamma^2 \| \bw \|_2^2.
\qedhere
\end{equation*}
\end{proof}
\end{proposition}

\section{Bayesian kernel width learning}
\label{sec:bayesian kernel width learning}

Section~\ref{sec:LinearLearnerRetrainingRidgeRegression} described how in a ridge regression setting, the marginal likelihood $\mathcal{L}(\lambda) = p( \by | \mathbf{X}, \lambda)$ can be efficiently computed for all $\lambda \in [0, \Lambda]$. With a prior $p(\lambda)$ over the lifetime (inverse kernel width) $\lambda$ whose support is included in $[0, \Lambda]$, the posterior distribution over $\lambda$ is
\begin{equation*}
p(\lambda | \by, \mathbf{X})
\propto p(\lambda) p( \by | \mathbf{X}, \lambda)
\end{equation*}
with normalizing constant
\begin{equation*}
p(\by | \mathbf{X})
= \sum_{c = 0}^{C-M} p( \by | \mathbf{X}, \lambda = \tau_c) \int_{\tau_c}^{\tau_{c + 1}} p(\lambda) \d \lambda
\end{equation*}
where $0 = \tau_0 < \tau_1 < \cdots < \tau_{C - M}$ is the sequence of times when new cuts appeared in any of the $M$ Mondrian samples. The predictive distribution at a new test point $\mathbf{x}_{*}$ is obtained by marginalizing out $\lambda$:
\begin{align*}
& p(y_{*} | \bx_{*}, \mathbf{X}, \by) \\
&=
  \int p(y_{*} | \bx_{*}, \mathbf{X}, \by, \lambda) p(\lambda | \by, \bx) \d \lambda
\\&=
  \sum_{c = 0}^{C - M} p(y_{*} | \bx_{*}, \lambda = \tau_c) p(\tau_c \leq \lambda < \tau_{c + 1} | \by, \mathbf{X})
\\&=
  \sum_{c = 0}^{C - M} p(y_{*} | \bx_{*}, \lambda = \tau_c) \int_{\tau_c}^{\tau_{c + 1}} p(\lambda | \by, \mathbf{X}) \d \lambda
\\&=
  \sum_{c = 0}^{C - M} p(y_{*} | \bx_{*}, \lambda = \tau_c) \int_{\tau_c}^{\tau_{c + 1}} \frac{p(\lambda) p( \by | \mathbf{X}, \lambda)}{p(\by | \mathbf{X})} \d \lambda
\\&=
  \sum_{c = 0}^{C - M} p(y_{*} | \bx_{*}, \lambda = \tau_c) p( \by | \mathbf{X}, \lambda = \tau_c) \frac{\int_{\tau_c}^{\tau_{c + 1}} p(\lambda) \d \lambda}{p(\by | \mathbf{X})}
\\&=
  \frac
  {\sum_{c = 0}^{C - M} p(y_{*} | \bx_{*}, \lambda = \tau_c) p( \by | \mathbf{X}, \lambda = \tau_c) \int_{\tau_c}^{\tau_{c + 1}} p(\lambda) \d \lambda}
  {\sum_{c = 0}^{C-M} p( \by | \mathbf{X}, \lambda = \tau_c) \int_{\tau_c}^{\tau_{c + 1}} p(\lambda) \d \lambda}
\\ &=
  \sum_{c = 0}^{C - M} k_c p(y_{*} | \bx_{*}, \lambda = \tau_c)
\end{align*}
where the mixing coefficients
\begin{equation*}
k_c
:= \frac
{p( \by | \mathbf{X}, \lambda = \tau_c) \int_{\tau_c}^{\tau_{c + 1}} p(\lambda) \d \lambda}
{\sum_{c = 0}^{C-M} p( \by | \mathbf{X}, \lambda = \tau_c) \int_{\tau_c}^{\tau_{c + 1}} p(\lambda) \d \lambda}
\end{equation*}
can be precomputed and cached for faster predictions. The integrals $\int_{\tau_c}^{\tau_{c + 1}} p(\lambda) \d \lambda$ can be readily evaluated if we have access to the cumulative distribution function of our prior $p(\lambda)$, which we assume.

\newpage

\section{Online learning}
\label{sec:SuppOnlineRidgeRegression}

Mirroring Section~\ref{sec:LinearLearnerRetrainingRidgeRegression}, we discuss the example of ridge regression where exact online updates can be carried out. Assume we have access to the regularized feature covariance matrix $\mathbf{A} = \bPhi^T \bPhi + \delta^2 \mathbf{I}_C$ and its inverse $\mathbf{A}^{-1}$ or Cholesky decomposition $\chol(\mathbf{A})$ before a new data point $\bx \in \Reals^D$ arrives, and we wish to update these efficiently.

If the dimensionality of $\phi$ increases by $k$ due to $\bx$ creating $k$ new non-empty partition cells, we first append $k$ rows and columns to $\mathbf{A}$, containing $0s$ only, except on the main diagonal we put $\delta^2$. Correspondingly, $\mathbf{A}^{-1}$ or $\chol(A)$ are updated by appending $k$ rows and columns, with non-zero entries only on the main diagonal. (These entries would equal $\delta^{-2}$ in $\mathbf{A}^{-1}$ and $\delta$ in $\chol(\mathbf{A})$). This ensures the feature map $\phi$ now incorporates all necessary features.

Noting that the $(i, j)$-entry of $\mathbf{A} - \delta^2 \mathbf{I}_C$ counts data points belonging to partition cells $i$ and $j$ at the same time (this can be non-zero only if $i$, $j$ correspond to different Mondrian samples), normalized by $1/M$, and that the $(i, j)$-entry of the outer product $\phi(\bx) \phi(\bx)^T$ is $1/M$ if the new data point $\bx$ falls into both cells $i$ and $j$, and $0$ otherwise, we see that
\begin{equation*}
\mathbf{A}_{\text{new}}
\gets \mathbf{A}_{\text{old}} + \phi(\bx) \phi(\bx)^T
\end{equation*}
is a rank-$1$ update. Therefore both $\mathbf{A}^{-1}$ and $\chol(\mathbf{A})$ can be updated efficiently in $\mathcal{O}(C^2)$ time and the new MAP weights $\hat{\bw}_{\text{new}} = \mathbf{A}_{\text{new}}^{-1} (\bPhi^T \by)$ in $\mathcal{O}(M C)$ by exploiting sparsity of $\phi(\bx)$. The determinant of the rank-1 updated matrix $\mathbf{A}_{\text{new}}$ can also be updated in $\mathcal{O}(C^2)$ time using the Matrix determinant lemma, or obtained directly from the Cholesky decomposition (as the squared product of its diagonal entries) in $\mathcal{O}(C)$ time, allowing the training marginal likelihood to be updated in $\mathcal{O}(N M + C^2)$.

\end{document}